\documentclass[10pt,twocolumn,twoside]{IEEEtran}
\usepackage{multirow}
\usepackage{amsfonts}

\usepackage{cite}

\ifCLASSINFOpdf
\usepackage[pdftex]{graphicx}
\else
\usepackage[dvips]{graphicx}
\fi
\usepackage{tabularx}
\newcolumntype{Y}{>{\centering\arraybackslash}X}

\usepackage{amsmath}
\usepackage{color}

\usepackage{algorithm}
\usepackage{algorithmic}

\usepackage{array}

\ifCLASSOPTIONcompsoc
\usepackage[caption=false,font=normalsize,labelfont=sf,textfont=sf]{subfig}
\else
\usepackage[caption=false,font=footnotesize]{subfig}
\fi

\usepackage{stfloats}
\usepackage{url}

\begin{document}
	%
	\title{SSDH: Semi-supervised Deep Hashing\\ for Large Scale Image Retrieval}
	%
	%
	%
	
	\author{
		Jian Zhang
		and Yuxin Peng
		\thanks{This work was supported by the National Natural Science Foundation of China under Grant 61771025, Grant 61371128, and Grant
			61532005.}
		\thanks{The authors are with the Institute of Computer Science and Technology, Peking University, Beijing 100871, China. Corresponding author: Yuxin Peng (e-mail: pengyuxin@pku.edu.cn).}
	}
	
	%
	%

	\markboth{IEEE TRANSACTIONS ON Circuits and Systems for Video Technology}%
	{IEEE TRANSACTIONS ON Circuits and Systems for Video Technology}
	%

	\IEEEpubid{\begin{minipage}{\textwidth}\ \\[12pt] \centering
			Copyright~\copyright~2017 IEEE. Personal use of this material is permitted. However, permission to use this material for \\any other purposes must be obtained from the IEEE by sending an email to pubs-permissions@ieee.org.
		\end{minipage}}


	\maketitle
	
	\begin{abstract}
		Hashing methods have been widely used for efficient similarity retrieval on large scale image database. Traditional hashing methods learn hash functions to generate binary codes from hand-crafted features, which achieve limited accuracy since the hand-crafted features cannot optimally represent the image content and preserve the semantic similarity. Recently, several deep hashing methods have shown better performance because the deep architectures generate more discriminative feature representations. However, these deep hashing methods are mainly designed for supervised scenarios, which only exploit the semantic similarity information, but ignore the underlying data structures. In this paper, we propose the semi-supervised deep hashing (SSDH) approach, to perform more effective hash function learning by simultaneously preserving semantic similarity and underlying data structures. The main contributions are as follows: (1) We propose a \textit{semi-supervised loss} to jointly minimize the empirical error on labeled data, as well as the embedding error on both labeled and unlabeled data, which can preserve the semantic similarity and capture the meaningful neighbors on the underlying data structures for effective hashing. (2) A \textit{semi-supervised deep hashing network} is designed to extensively exploit both labeled and unlabeled data, in which we propose an \textit{online graph construction} method to benefit from the evolving deep features during training to better capture semantic neighbors. To the best of our knowledge, the proposed deep network is the \textit{first} deep hashing method that can perform hash code learning and feature learning simultaneously in a semi-supervised fashion. Experimental results on 5 widely-used datasets show that our proposed approach outperforms the state-of-the-art hashing methods.
	\end{abstract}
	
	
	%
	\IEEEpeerreviewmaketitle

	\section{Introduction}
	%
	%
	%
	%
	
	\IEEEPARstart{M}{ultimedia} data including images and videos have been growing explosively on the Internet in recent years, and retrieving similar data from a large scale database has become an urgent need. Hashing methods can map similar data to similar binary codes that have small Hamming distance, due to the low storage cost and fast retrieval speed, hashing methods have been receiving broad attention~\cite{imghashsurvey}. Hashing methods have also been used in many applications, such as image retrieval~\cite{gong2011iterative,irie2014locally,kan2014semisupervised,liu2011hashing,weiss2009spectral,7298947,xia2014supervised,zhao2015deep,zhu2016deep} and video retrieval~\cite{7937842,7855682,7851077}. Traditional hashing methods~\cite{gionis1999similarity,7298947,wang2010sequential,weiss2009spectral} take pre-extracted image features as input, and then learn hash functions by exploiting the data structures or applying the similarity preserving regularizations. These methods can be divided into two categories: unsupervised and supervised methods. Recently, inspired by the remarkable progress of deep networks, some deep hashing methods have also been proposed~\cite{7298947,xia2014supervised,zhang2015bit,zhao2015deep,zhu2016deep,Liu_2016_CVPR}.
	\IEEEpubidadjcol
	Unsupervised methods design hash functions by using unlabeled data to generate binary codes. Locality Sensitive Hashing (LSH)~\cite{gionis1999similarity} is a representative unsupervised method, which is proposed to use random linear projections to map data into binary codes. Instead of using randomly generated hash functions, some data-dependent methods~\cite{gong2011iterative,irie2014locally,liu2011hashing,weiss2009spectral,zhang2013topology} have been proposed to capture the data properties, such as data distributions and manifold structures. For example, Weiss \textit{et al.}~\cite{weiss2009spectral} propose Spectral Hashing (SH), which tries to keep hash functions balanced and uncorrelated. Liu \textit{et al.}~\cite{liu2011hashing} propose Anchor Graph Hashing (AGH) to preserve the neighborhood structures by anchor graphs, and Locally Linear Hashing (LLH)~\cite{irie2014locally} is proposed to use locality sensitive sparse coding to capture the local linear structures and then recover these structures in Hamming space. Gong \textit{et al.}~\cite{gong2011iterative} propose Iterative Quantization (ITQ) to generate hash codes by minimizing the quantization error of mapping data to the vertices of a binary hypercube. Topology Preserving Hashing (TPH)~\cite{zhang2013topology} is proposed to perform hash function learning by preserving consistent neighborhood rankings of data points in Hamming space. Shen \textit{et al.}~\cite{AIBC} propose Asymmetric Inner-product Binary Coding (AIBC) method, where two asymmetric coding functions are learned such that the inner products between original data pairs are approximated by the produced binary code vectors.
	
	While unsupervised methods are promising to retrieve the neighbors under some kind of distance metrics (such as $L_2$ distance), the neighbors in the feature space can not optimally reflect the semantic similarity. Therefore, supervised hashing methods~\cite{kulis2009learning,kulis2009fast,wang2010sequential,norouzi2011minimal,liu2012supervised,norouzi2012hamming,wang2013learning,wang2013order,li2013learning,lin2014fast,wang2015ranking,7298598} are proposed to utilize the semantic information such as image labels to generate effective hash codes. For example, Kulis \textit{et al.}~\cite{kulis2009learning} propose Binary Reconstruction Embedding (BRE) to learn hash functions by minimizing the reconstruction error between original distances and reconstructed distances in the Hamming space. Liu \textit{et al.}~\cite{liu2012supervised} propose Supervised Hashing with Kernels (KSH) to learn hash functions by preserving the pairwise relations between data samples provided by labels.  Wang \textit{et al.}~\cite{wang2010sequential} propose Semi-supervised Hashing (SSH) to learn hash functions by minimizing the empirical error over labeled data while maximizing the information entropy of the generated hash codes over both labeled and unlabeled data. Norouzi \textit{et al.}~\cite{norouzi2012hamming} propose to learn hash functions based on a triplet ranking loss that can preserve relative semantic similarity. Ranking Preserving Hashing (RPH)~\cite{wang2015ranking} and Order Preserving Hashing (OPH)~\cite{wang2013order} are proposed to learn hash functions by preserving the ranking information, which is obtained based on the number of shared semantic labels between data examples. Shen \textit{et al.}~\cite{7298598} propose Supervised Discrete Hashing (SDH) to leverage label information to obtain hash codes by integrating hash code generation and classifier training. Shen \textit{et al.}~\cite{DPLM} further improve SDH by a discrete proximal linearized minimization method, which directly handles the discrete constraints during the learning process and improves the performance.
	
	For most of the unsupervised and supervised hashing methods, input images are represented by hand-crafted features (e.g. GIST~\cite{oliva2001modeling}), which can not optimally represent the semantic information of images. Inspired by the fast progress of deep networks on image classification~\cite{krizhevsky2012imagenet}, some deep hashing methods have been proposed~\cite{7298947,xia2014supervised,zhang2015bit,zhao2015deep,zhu2016deep,Liu_2016_CVPR} to take advantage of the superior feature representation power of deep networks. Convolutional Neural Network Hashing (CNNH)~\cite{xia2014supervised} is a two-stage framework based on the convolutional networks, which learns fixed hash codes in the first stage, and learns hash functions and image representations in the second stage. Although the learned hash codes can guide feature learning, the learned image features cannot provide feedback for learning better hash codes. To overcome the shortcomings of the two-stage learning scheme, some approaches have been proposed to perform simultaneously feature learning and hash code learning. Network in Network Hashing (NINH)~\cite{7298947} is proposed to design a triplet ranking loss to capture the relative similarities of images. NINH is a one-stage supervised method, thus image representation learning and hash code learning can benefit each other in the deep architecture. Some similar ranking-based deep hashing methods~\cite{zhang2015bit,zhao2015deep} have also been proposed recently, which also design hash functions to preserve the ranking information obtained by labels.
	
	However, existing deep hashing methods~\cite{7298947,xia2014supervised,zhang2015bit,zhao2015deep,zhu2016deep,Liu_2016_CVPR} are mainly supervised methods. On one hand, they heavily rely on labeled images and require a large amount of labeled data to achieve better performance. But labeling images consumes large amounts of time and human labors, which is not practical in real world applications. One the other hand, they only consider semantic information while ignore the underlying data structures of unlabeled data. Thus it is necessary to make full use of unlabeled images to improve the deep hashing performance. In this paper, we propose the semi-supervised deep hashing (SSDH) approach, which learns hash functions by preserving the semantic similarity and underlying data structures simultaneously. To the best of our knowledge, the proposed SSDH is the first deep hashing method that can perform hash code learning and feature learning simultaneously in a semi-supervised fashion. The main contributions of this paper can be concluded as follows:
	\begin{itemize}
		\item \textit{Semi-supervised loss}. Existing deep hashing methods only design the loss functions to preserve the semantic similarity but ignore the underlying data structures, which is essential to capture the semantic neighborhoods and return the meaningful nearest neighbors for effective hashing~\cite{liu2011hashing}. To address the problem, we propose a semi-supervised loss function to exploit the underlying structures of unlabeled data for more effective hashing. By jointly minimizing the empirical error on labeled data as well as the embedding error on both labeled and unlabeled data, the learned hash functions can not only preserve the semantic similarity, but also capture the meaningful neighbors on the underlying data structures. Thus the proposed SSDH method can benefit greatly from both labeled and unlabeled data.
		\item \textit{Semi-supervised deep hashing network}. A deep network structure is designed to learn hash functions and image representations simultaneously in a semi-supervised fashion. The proposed deep network consists of three pars: representation learning layers extract discriminative deep features from the input images, hash code learning layer maps the image features into binary hash codes and classification layer predicts the pseudo labels of unlabeled data. Finally the proposed semi-supervised loss function is designed to model the semantic similarity constraints, meanwhile preserve the underlying data structures of image features. The whole network is trained in an end-to-end way, thus our proposed method can perform hash code and feature learning simultaneously.
		\item \textit{Online graph construction}. The traditional offline graph construction methods are time consuming due to the $O(n^2)$ complexity, which is intractable for large scale data. To address this issue, we propose an online graph construction strategy. Rather than constructing neighborhood graph over all the data in advance, our online graph construction strategy constructs the neighborhood graph in a mini-batch during the training procedure. On one hand, online graph construction only needs to take consideration of a much smaller mini-batch of data, which is efficient and suitable for the batch-based deep network's training. On the other hand, online graph construction can benefit from the evolving feature representations extracted from deep networks.
	\end{itemize}
	
	Experiments on 5 widely-used datasets show that our proposed SSDH approach achieves the best search accuracy comparing with 8 state-of-art methods. The rest of this paper is organized as follows. Section II presents a brief review of related deep hashing methods. Section II introduces our proposed SSDH approach. Section IV shows the experiments on the 5 widely-used image datasets. Finally Section V concludes this paper.
	
	\section{Related Work}
	Hashing methods have become one of the most popular methods for similarity retrieval on large scale image datasets. In recent years, some hashing methods have been proposed to perform hash function learning with the deep networks and have made considerable progress. In this section, we briefly review some representative deep hashing methods.
	\subsection{Convolutional Neural Network Hashing}
	Convolutional Neural Network Hashing (CNNH)~\cite{xia2014supervised} is a two-stage framework, including a hash code learning stage and a hash function learning stage. Given the training images $\mathcal{I} = \{I_1,I_2,\ldots,I_n\}$, in the hash code learning stage (Stage 1), CNNH learns an approximate hash code for each training data by optimizing the following loss function:
	\begin{equation}
		\label{cnnh_obj_1}
		\min \limits_H {\|S-\frac{1}{q}HH^T\|^2_F}
	\end{equation}
	where $\|\cdot\|_F$ denotes the Frobenius norm. $S$ denotes the semantic similarity of the image pairs in $\mathcal{I}$, in which $S_{ij}=1$ when image $I_i$ and $I_j$ are semantically similar, otherwise $S_{ij}=-1$. $H \in \{-1,1\}^{n \times q}$ denotes the approximate hash codes. For the training images, $H$ encodes the approximate hash codes which preserve the pairwise similarities in $S$. It's difficult to directly optimize equation~(\ref{cnnh_obj_1}), CNNH firstly relaxes the integer constraints on $H$ and randomly initializes $H \in [-1,1]^{n \times q}$, then optimizes equation~(\ref{cnnh_obj_1}) using a coordinate descent algorithm, which sequentially chooses one entry in $H$ to update while keeping other entries fixed. Thus it is equivalent to optimize the following equation:
	\begin{equation}
		\min \limits_{H_{\cdot j}}{\|H_{\cdot j}H_{\cdot j}^T-(qS-\sum \limits _{c \not = j}{H_{\cdot c}H_{\cdot c}^T})\|^2_F}
	\end{equation}
	where $H_{\cdot j}$ and $H_{\cdot c}$ denote the $j$-th and the $c$-th column of $H$ respectively. In the hash function learning stage (Stage 2), CNNH utilizes deep networks to learn the image feature representation and hash functions. Specifically, CNNH adopts the deep framework in~\cite{hinton2012improving} as its basic network, and designs an output layer with sigmoid activation to generate $q$-bit hash codes. CNNH trains the designed deep network in a supervised way, in which the approximate hash codes learned in Stage 1 are used as the ground-truth. In addition, if discrete class labels of training images are available, CNNH also incorporates these image labels to hash functions.
	
	Based on the deep network, CNNH learns deep features and hash functions at the same time. However CNNH is a two-stage framework, where the learned deep features in Stage 2 cannot help to improve the approximate hash code learning in Stage 1, which limits the performance of hash function learning.
	
	\subsection{Network in Network Hashing}
	Different from the two-stage framework in CNNH~\cite{xia2014supervised}, Network in Network Hashing (NINH)~\cite{7298947} integrates image representation learning and hash code learning into a one-stage framework. NINH constructs the deep framework for hash function learning based on the Network in Network architecture~\cite{lin2013network}, with a shared sub-network composed of several stacked convolutional layers to extract image features, as well as a divide-and-encode module encouraged by sigmoid activation function and a piece-wise threshold function to output binary hash codes. During the learning process, without generating approximate hash codes in advance, NINH designs a triplet ranking loss function to exploit the relative similarity of the training images to directly guide hash function learning:
	\begin{equation}
		\small
		\label{ninh_loss}
		\begin{split}
			& l_{triplet}(I,I^+,I^-) = \\
			& \max(0,1-(\|b(I)-b(I^-)\|_{\mathcal{H}}-\|b(I)-b(I^+)\|_{\mathcal{H}})) \\
			& s.t. \quad b(I), b(I^+), b(I^-) \in \{-1,1\}^q
		\end{split}
	\end{equation}
	where $I,I^+$ and $I^-$ specify the triplet constraint that image $I$ is more similar to image $I^+$ than to image $I^-$ based on image labels, $b(\cdot)$ denotes binary hash code, and $\|\cdot\|_\mathcal{H}$ denotes the Hamming distance. For simplifying optimization of equation~(\ref{ninh_loss}), NINH applies two relaxation tricks: relaxing the integer constraint on binary hash code, and replacing Hamming distance with Euclidean distance.
	\begin{figure*}[!ht]
		\centering
		\includegraphics[width=0.8\textwidth]{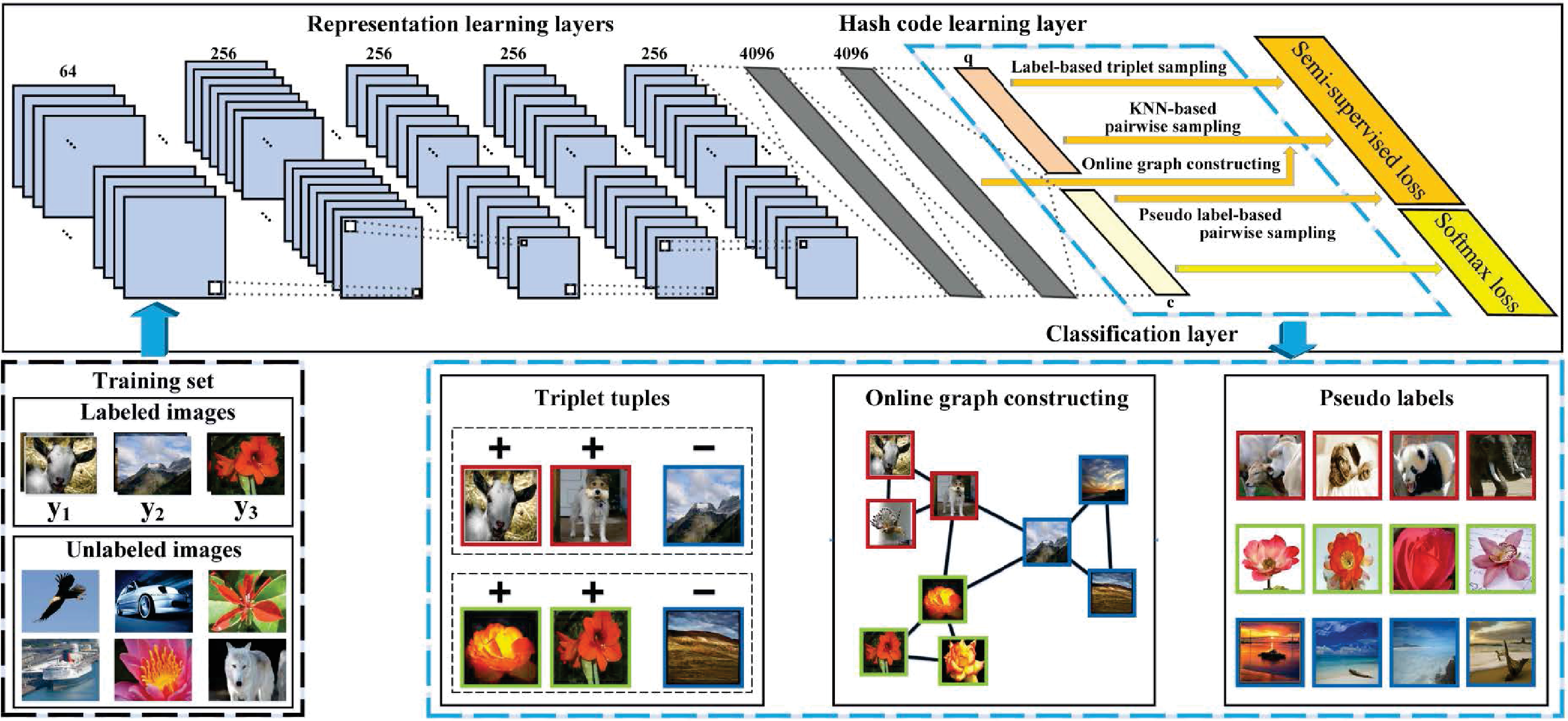}
		\caption{Overview of the proposed SSDH framework, which consists of three parts: representation learning layers, hash code learning layer and classification layer. Finally a semi-supervised loss is proposed to guide the network training.}
		\label{framework}
	\end{figure*}
	\subsection{Bit-scalable Deep Hashing}
	The Bit-scalable Deep Hashing (BS-DRSCH) method~\cite{zhang2015bit} is also a one-stage deep hashing method that constructs an end-to-end architecture for hash function learning. In addition, BS-DRSCH is able to flexibly manipulate hash code length by weighing each bit of hash codes. BS-DRSCH defines a weighted Hamming affinity to measure the dissimilarity between two hash codes:
	\begin{equation}
		\mathcal{H}(b(I_i),b(I_j)) = -\sum \limits_{k = 1}^q w_k^2b_k(I_i)b_k(I_j)
	\end{equation}
	where $I_i$ and $I_j$ are training images, $b(\cdot)$ denotes the $q$-bit binary hash code, and $b_k(\cdot)$ denotes the $k$-th hash bit in $b(\cdot)$, and $w_k$ is the weight value attached to $b_k(\cdot)$, which is to be learned and expected to represent the significance of $b_k(\cdot)$. Similar to NINH~\cite{7298947}, BS-DRSCH organizes the training images into triplet tuples, then formulates the loss function as follows:
	\begin{equation}
		\small
		\label{bit_scale_loss}
		\centering
		\begin{split}
			loss & = \sum \limits_{I, I^+, I^-} \max(\mathcal{H}(b(I),b(I^+))-\mathcal{H}(b(I),b(I^-)), C) \\
			& + \frac{1}{2}\sum \limits_{i,j}\mathcal{H}(b(I_i),b(I_j))S_{ij} 
		\end{split}
	\end{equation}
	where $S_{ij}$ denotes the similarity between images $I_i$ and $I_j$ based on labels. The defined $loss$ is the sum of two terms: the first term is a max-margin term, which is designed to preserve the ranking orders. The second term is a regularization term, which is defined to keep the pairwise adjacency relations. BS-DRSCH builds a deep convolutional neural network to perform hash function learning, in which the convolutional layers are used to extract deep features, while the fully-connected layer followed by the tanh-like layer is used to generate hash codes. For learning the weight values $\{w_k\}, k=1,\ldots,q$, BS-DRSCH designs an element-wise layer on the top of the deep architecture, which is able to weigh each bit of hash code.
	
	Both NINH~\cite{7298947} and BS-DRSCH~\cite{zhang2015bit} are supervised deep hashing methods, and they only preserve the semantic similarity based on labeled training data, but ignore the underlying data structures of unlabeled data, which is essential to capture the semantic neighborhoods for effective hashing.
	
	\section{Semi-supervised deep hashing}
	\subsection{Overview of proposed hashing framework}
	Given a set of $n$ images $\mathcal{X}$ that are composed of labeled images $(\mathcal{X}_L,\mathcal{Y}_L)=\{(x_1,y_1),(x_2,y_2),\ldots,(x_l,y_l)\}$ and unlabeled images $\mathcal{X}_U = \{x_{l+1},x_{l+2},\ldots, x_n\}$, where $\mathcal{Y}_L$ is the label information. The goal of hashing methods is to learn a mapping function $\mathcal{B}:X \rightarrow \{-1,1\}^q$, which encodes an image $X \in \mathcal{X}$ into a \textit{q}-bit binary codes $\mathcal{B}(X)$ in the Hamming space, while preserving the semantic similarity of images.
	
	In this paper, we propose a novel deep hashing method called semi-supervised deep hashing (SSDH), to learn better hash codes by preserving the semantic similarity and underlying data structures simultaneously. As shown in Fig.~\ref{framework}, the deep architecture of SSDH includes three main components: 1) A deep convolutional neural network is designed to learn and extract discriminative deep features for images, which takes both the labeled and unlabeled image data as input. 2) A hash code learning layer is designed to map the image features into \textit{q}-bit hash codes. 3) A semi-supervised loss is designed to preserve the semantic similarity, as well as the neighborhood structures for effective hashing, which is formulated as minimizing the empirical error on the labeled data as well as the embedding error on both the labeled and unlabeled data. The above three components are coupled into a unified framework, thus our proposed SSDH can perform image representation learning and hash code learning simultaneously. In the following of this section, we first introduce the proposed deep network architecture, then we introduce the proposed semi-supervised loss function as well as the training of the proposed network.
	
	\subsection{The Proposed Deep Network}
	We adopt the Fast CNN architecture (denoted as CNN-F for simplification) from~\cite{chatfield2014return} as the base network. The first seven layers are configured the same as those in CNN-F~\cite{chatfield2014return}, which build the representation learning layers to extract discriminative image features. The eighth layer is the hash code learning layer, which is constructed by a fully-connected layer followed by a sigmoid activation function, and its outputs are hash codes defined as:
	\begin{equation}
		h(x) = sigmoid(W^Tf(x)+v)
	\end{equation}
	where $f(x)$ is deep feature extracted from the last representation learning layer, i.e., the output of fully-connected layer fc7, $W$ denotes the weights in the hash code learning layer, and $v$ is the bias parameter. Through the hash code learning layer, image features $f(x)$ are mapped into $[0,1]^q$. Since hash codes $h(x) \in [0,1]^q$ are continuous real values, we apply a thresholding function to obtain binary codes:
	\begin{equation}
		\label{binarycode}
		b_k(x) = sgn(h_k(x)-0.5), \quad k=1,2,\cdots,q
	\end{equation}
	However, binary codes are hard to directly optimize, thus we relax binary codes $b(x)$ with continuous real valued hash codes $h(x)$ in the rest of this paper.
	
	The ninth layer is the classification layer, which predicts the pseudo labels of unlabeled data, we use the cross-entropy loss function to train the classification layer. Outputs of the last representation learning layer (fc7), the hash code learning layer (fc8) and the classification layer (fc9) are connected with the semi-supervised loss function, so that the proposed loss function can make full use of the underlying data structures of extracted image features and pseudo labels to improve hash function learning performance. The whole network is an end-to-end structure, which can perform hash function learning and feature learning simultaneously.
	\begin{table}[h]
		\centering
		\caption{Detailed network configurations of our proposed SSDH approach}
		\label{Network configuration}
		\begin{tabularx}{0.40\textwidth}{c|Y}
			\hline
			Layers & Configuration \\ \hline
			conv1 & filter 64x11x11, st. 4x4, pad 0, LRN, pool 2x2 \\ \hline
			conv2 & filter 256x5x5, st. 1x1, pad 2, LRN, pool 2x2\\ \hline
			conv3 & filter 256x3x3, st. 1x1, pad 1 \\ \hline
			conv4 &  filter 256x3x3, st. 1x1, pad 1\\ \hline
			conv5 &  filter 256x3x3, st. 1x1, pad 1, pool 2x2\\ \hline
			fc6 &  4096, dropout\\ \hline
			fc7 &  4096, dropout\\ \hline
			fc8 &  \textit{q}-dimensional, sigmoid\\ \hline
			fc9 &  \textit{c}-dimensional \\ \hline
		\end{tabularx}
	\end{table}
	
	The detailed configurations of our network structure are shown in Table~\ref{Network configuration}. For the five convolutional layers: the ``filter'' parameter specifies the number and the receptive filed size of convolutional kernels as $num \times size \times size$, the ``st.'' and ``pad'' parameters specify the convolution stride and spatial padding respectively, the ``LRN'' indicates whether Local Response Normalization (LRN)~\cite{krizhevsky2012imagenet} is used, the ``pool'' indicates whether to apply the max-pooling downsampling, and specifies the pooling window size by $size \times size$. For the fully-connected layers, we specify their dimensions, of which the first two are 4096, the hash layer is the same as hash code length \textit{q}, and the classification layer is the same as the class number \textit{c}. The ``dropout'' indicates whether the fully-connected layer is regularized by dropout~\cite{krizhevsky2012imagenet}. The hash code learning layer takes sigmoid as activation function, while the other weighted layers take Rectiﬁed Linear Unit (ReLU)~\cite{krizhevsky2012imagenet} as activation functions. It is noted that although we adopt CNN-F~\cite{chatfield2014return} as our base network, it's practicable to replace the base network by other deep convolutional network structures, such as Network in Network (NIN)~\cite{lin2013network} model or AlexNet~\cite{krizhevsky2012imagenet} model.
	
	\subsection{Semi-supervised Deep Hashing Learning}
	The proposed semi-supervised loss function is composed of three terms, namely supervised ranking term, semi-supervised embedding term, and pseudo-label term. The first term can preserve the semantic similarity by minimizing the empirical error. The second term can capture the underlying data structures by minimizing the embedding error, which measures whether hash functions preserve the local structure of image data when embedding images into Hamming space, namely the neighbor points should have similar hash codes while non-neighbor points should have dissimilar hash codes. The third term encourages the decision boundary in low density regions, which is complementary to semi-supervised embedding term. Thus the proposed SSDH can fully benefit from both labeled and unlabeled data. We introduce these three terms separately in the following parts of this subsection.
	\subsubsection{Supervised Ranking Term}
	\label{rankingterm_sec}
	Encouraged by \cite{7298947}, we formulate the supervised ranking term as triplet ranking regularization to preserve the relative semantic similarity provided by label annotations. For labeled images $(\mathcal{X}_L,\mathcal{Y}_L)$, we randomly sample a set of triplet tuples depending on the labels, $\mathcal{T}=\{(x_i^L,x_i^{L^+},x_i^{L^-})\}^t_{i=1}$, in which $x_i^L$ and $x_i^{L^+}$ are two similar images with the same labels, while $x_i^L$ and $x_i^{L^-}$ are two dissimilar images with different labels, and $t$ is the number of sampled triplet tuples. For the randomly sampled triplet tuples $(x_i^L,x_i^{L^+},x_i^{L^-}), i=1 \cdots t$, the supervised ranking term is defined as:	
	\begin{equation}
		\small
		\label{rankingterm2}
		\begin{split}
			&\mathcal{J}_L(x_i^L,x_i^{L^+},x_i^{L^-}) = \\
			&\max(0, m_t+\|h(x_i^L)-h(x_i^{L^+})\|^2-\|h(x_i^L)-h(x_i^{L^-})\|^2)
		\end{split}
	\end{equation}
	where $\|\cdot\|^2$ denotes the Euclidean distance, and the constant parameter $m_t$ defines the margins between the relative similarity of the two pairs $(x_i^L,x_i^{L^+})$ and $(x_i^L,x_i^{L^-})$, that is to say, we expect the distance of dissimilar pair $(x_i^L,x_i^{L^-})$ to be larger than the distance of similar pair $(x_i^L,x_i^{L^+})$ by at least $m_t$. Thus minimizing $\mathcal{J}_L$ can reach our goal to preserve the semantic ranking information provided by labels.
	
	\subsubsection{Semi-supervised Embedding Term}
	Although we can train the network solely based on supervised ranking term, the deep models with multiple layers often require a large amount of labeled data to achieve better performance. Usually it's hard to collect abundant labeled images, since labeling images consumes large amounts of time and human labors. Thus it's necessary to improve search accuracy by utilizing unlabeled data which are much easier to obtain. Manifold assumption~\cite{chapelle2006semi} states that data points within the same manifold are likely to have the same labels. Based on this assumption, the graph-based embedding algorithms are often used to capture the underlying manifold structures of data. Thus we add an embedding term to capture the underlying structures of both labeled and unlabeled data.
	
	Modeling the neighborhood structures often requires to build a neighborhood graph. A neighborhood graph is defined as $G = (V,E)$, where the vertices \textit{V} represent the data points, and the edges \textit{E}, which are often represented by an $n\times n$ adjacency matrix \textit{A}, indicate the adjacency relations among the data points in a specified space (usually in the feature space). But for large scale data, building neighborhood graph on all the data is time and memory consuming, since the time complexity and memory cost of building the \textit{k}-NN graph on all the data are both $O(n^2)$, which is intractable for large \textit{n}. Instead of constructing graph offline, we propose to construct graph online within a mini-batch by using the features extracted from the proposed deep networks. More specifically, given the mini-batch including $r$ images, $\mathcal{X}^B=\{x_i\}_{i=1}^r$, and the deep features $\{f(x_i): x_i \in \mathcal{X}^B\}$. We firstly use deep features to calculate the \textit{k-nearest-neighbors} $NN_k(i)$ for each image $x_i$, then we can construct the adjacency matrix as:
	\begin{equation}
		\small
		\label{equknn}
		A(i,j)=\left\{ \begin{array}{rrr}
			1 :&   I(i,j) = 0 \land x_j \in NN_k(i)\\
			0 :&  I(i,j) = 0 \land x_j \notin NN_k(i) \\
			-1 :&   I(i,j) = 1
		\end{array}
		\right.
	\end{equation}
	where $I(i,j)$ is an indicator function: $I(i,j)=1$ if $x_i$ and $x_j$ are both labeled images, otherwise $I(i,j)=0$. $A(i,j)=-1$ means that we don't build neighborhood graph between two labeled images, since we can use supervised ranking term to model their similarity relations. But we construct neighborhood graph between labeled and unlabeled images, so that we can make unlabeled neighbors of labeled images close to them and improve the retrieval accuracy. Thus the embedding term is in a semi-supervised fashion. Because the size of mini-batch is much smaller than the size of the whole training data, i.e., $r \ll n$, thus it is efficient to compute online in each iteration. Furthermore, building neighborhood graph online can benefit from the representative deep features, since the training procedure of deep network generates more and more discriminative feature representations, which can capture the semantic neighbors more accurately.
	
	Based on the constructed adjacency matrix, we expect the neighbor image pairs to have small distances, while the non-neighbor image pairs to have large distances. We apply pairwise contrastive regularization~\cite{hadsell2006dimensionality} to achieve this goal. Thus for one image pair $(x_i,x_j)$, our semi-supervised embedding term is formulated as:
	\begin{equation}
		\small
		\label{embeddingterm2}
		\begin{split}
			&\mathcal{J}_U(x_i,x_j)=\\
			&\left\{ \begin{array}{cc}
				\|h(x_i)-h(x_j)\|^2   & , A(i,j)=1 \\
				\max(0,m_p-\|h(x_i)-h(x_j)\|^2)   & , A(i,j)=0
			\end{array}
			\right.
		\end{split}
	\end{equation}
	where $\mathcal{J}_U(x_i,x_j)$ is the contrastive loss between the two images $x_i$ and $x_j$ based on the neighborhood graph, $\|\cdot\|^2$ represents the Euclidean distance, and the constant $m_p$ is a margin parameter for the distance metric of non-neighbor image pairs. With the neighborhood graphs constructed on the mini-batches, the semi-supervised embedding term $\mathcal{J}_U$ captures the embedding error on both labeled and unlabeled data. We minimize $\mathcal{J}_U$ to preserve the underlying neighborhood structures of image data.
	
	\subsubsection{Pseudo-label Term}
	Apart from proposed graph approach, Lee \textit{et al.}~\cite{lee2013pseudo} propose the pseudo-label method for image classification task, which can also make use of unlabeled data. Pseudo-label method predicts unlabeled data with the class that has the maximum predicted probability as their labels. The maximum a posteriori estimation encourages this approach to have the decision boundary in low density regions, which improves the generalization of classification model. Different from our proposed graph approach, which encourages neighborhood image pairs to have small distance and non-neighbor image pairs to have large distance, pseudo-label method encourages each image data to have an 1-of-C prediction, these two approaches are complementary to each other. Thus we integrate pseudo-label method in our framework. More specifically, in our network structure, the top layers have two branches: one is hash code learning layer, and the other is a classification layer, which uses softmax function to calculate the label probability of each unlabeled data.
	\begin{equation}
		p_j(i)=\dfrac{e^{z_{ij}}}{\sum_{l=1}^{c}e^{z_{il}}}
	\end{equation}
	where $z_{i\cdot}$ is the output of fc9 layer for image $x_i$, and $c$ is the class number. During the forward propagation of the model training, the fc9 layer in the network structure predicts the pseudo-labels, and its parameters are from previous iteration. We also take the class with maximum predicted probability as their true label. Then we also use the contrastive loss function to model semantic similarity between the pseudo labels. Similar to equation~(\ref{embeddingterm2}), the pseudo-label term is defined as follows:
	\begin{equation}
		\small
		\label{pseudoterm}
		\begin{split}
			&\mathcal{J}_P(x_i,x_j)=\\
			&\left\{ \begin{array}{cc}
				\|h(x_i)-h(x_j)\|^2   & , PL_i=PL_j \\
				\max(0,m_p-\|h(x_i)-h(x_j)\|^2)   & , PL_i\neq PL_j
			\end{array}
			\right.
		\end{split}
	\end{equation}
	where the $PL_i$ and $PL_j$ are the predicted pseudo labels. For the data that have multiple labels, we follow the idea of ML-KNN~\cite{MLKNN} to calculate the $k$-nearest labeled neighbor of each unlabeled data $x$ and assign the labels of neighbors to $x$. If $x_i$ and $x_j$ do not share any labels, then $PL_i\neq PL_j$. If the predicted pseudo labels with maximum predicted probability are the same, then $PL_i=PL_j$.
	
	\subsubsection{The Semi-supervised Loss}
	Combining the supervised ranking term, semi-supervised embedding term and pseudo-label term, we define the semi-supervised loss function as:
	\begin{equation}
	\small
		\begin{split}
			& \mathcal{J} = \sum_{i=1}^t{\mathcal{J}_L(x_i^L,x_i^{L^+},x_i^{L^-})}+\lambda\sum_{i,j=1}^n{\mathcal{J}_U(x_i,x_j)} \\
			& +\mu\sum_{i=1}^n{\mathcal{J}_P(x_i,x_j)} \\
			& s.t. \quad x_i^L,x_i^{L^+},x_i^{L^-} \in \mathcal{X_L}, \quad x_i,x_j \in \mathcal{X}
		\end{split}
	\end{equation}
	where $t$ is the number of triplet tuples sampled from labeled images, which is the same as the number of labeled images in the mini-batch, that is to say we randomly generate one triplet tuple for each labeled image. $\lambda$ and $\mu$ are the balance parameters. In order to make the neighbor and non-neighbor image pairs to be balanced, we only use a small part of image pairs obtained by the neighborhood graph to calculate the loss. More specifically, for each image $x_i$, we randomly select a neighbor pair $(x_i,x_i^+)$, and a non-neighbor pair $(x_i,x_i^-)$ for semi-supervised embedding term, and we randomly select a positive pair $(x_i^L,x_i^{L^+_p})$ and negative pair $(x_i^L,x_i^{L^-_p})$ for pseudo-label term, then we rewrite the loss function as:
	\begin{equation}
		\small
		\label{jointloss}
		\begin{split}
			& \mathcal{J} = \sum_{i}^t{\mathcal{J}_L(x_i^L,x_i^{L^+},x_i^{L^-})}+\lambda\sum_{i=1}^n(\mathcal{J}_U(x_i,x_i^+)+\mathcal{J}_U(x_,x_i^-)) \\
			& +\mu\sum_{i=1}^n(\mathcal{J}_P(x_i^L,x_i^{L^+_p})+\mathcal{J}_P(x_i^L,x_i^{L^-_p}))
		\end{split}
	\end{equation}
	
	\subsection{Network Training}
	The proposed SSDH network is trained in a mini-batch way, that is to say, we only use a mini-batch of training images to update the network's parameters in one iteration. For each iteration, the input image batch $\mathcal{X}^B \subset \mathcal{X}$ is composed of some labeled images $(\mathcal{X}_L^B,\mathcal{Y}_L^B)=\{(x_1,y_1),(x_2,y_2),\cdots,(x_{m},y_{m})\}$ and some unlabeled images $\mathcal{X}_U^B = \{x_{{m}+1},x_{{m}+2},\cdots, x_r\}$, where ${m}$ denotes the number of labeled images in the mini-batch, and ${r}$ denotes the total number of images in the mini-batch.
	
	The proposed network structure has two branches, namely the hash code learning layer and the classification layer. The hash code learning layer generates the hash codes while the classification layer predicts the pseudo-labels of unlabeled data. We use a joint training scheme to learn the network simultaneously. For the classification layer, we use labeled data to train it with a cross-entropy loss, which can make the predicted pseudo labels more accurate. For the hash code learning layer, with the equations (\ref{rankingterm2}), (\ref{embeddingterm2}), (\ref{pseudoterm}) and (\ref{jointloss}), we adopt stochastic gradient descent (SGD) to train the deep network. For each triplet tuple and image pair, we use the back-propagation algorithm (BP) to update the parameters of the network. More specifically, according to equation (\ref{rankingterm2}), we can compute the sub-gradient for each triplet tuple $(x_i^L,x_i^{L^+},x_i^{L^-})$, with respect to $h(x_i^L)$, $h(x_i^{L^+})$ and $h(x_i^{L^-})$ respectively as:
	\begin{equation}
		\small
		\begin{split}
			&\frac{\partial{\mathcal{J}_L}}{\partial{h(x_i^L)}} = 2(h(x_i^{L^-})-h(x_i^{L^+}))\times I_{c} \\
			&\frac{\partial{\mathcal{J}_L}}{\partial{h(x_i^{L^+})}} = 2(h(x_i^{L^+})-h(x_i^L))\times I_{c} \\
			&\frac{\partial{\mathcal{J}_L}}{\partial{h(x_i^{L^-})}} = 2(h(x_i^L)-h(x_i^{L^-}))\times I_{c} \\
			& I_{c} = I_{m_t+\|h(x_i^L)-h(x_i^{L^+})\|^2-\|h(x_i^L)-h(x_i^{L^-})\|^2>0}
		\end{split}
	\end{equation}
	where $I_{c}$ is an indicator function, $I_{c} = 1$ if $c$ is true, otherwise $I_{c} = 0$.
	
	For the semi-supervised embedding term and pseudo-label term, according to equations (\ref{embeddingterm2}) and (\ref{pseudoterm}), when $A(i,j)=1$ or $PL_i=PL_j$, we compute the sub-gradient for $h(x_i)$ and $h(x_j)$ respectively as:
	\begin{equation}
	\small
		\begin{split}
			\frac{\partial{\mathcal{J}_U}}{\partial{h(x_i)}} = 2(h(x_i)-h(x_j)) \\
			\frac{\partial{\mathcal{J}_U}}{\partial{h(x_j)}} = 2(h(x_j)-h(x_i))
		\end{split}
	\end{equation}
	and when $A(i,j)=0$ or $PL_i\neq PL_j$, we compute the sub-gradient for $h(x_i)$ and $h(x_j)$ respectively as:
	\begin{equation}
			\small
		\begin{split}
			&\frac{\partial{\mathcal{J}_U}}{\partial{h(x_i)}} = 2(\|h(x_i)-h(x_j)\|-m_p)\times sgn(h(x_i)-h(x_j))\times I_c \\
			&\frac{\partial{\mathcal{J}_U}}{\partial{h(x_j)}} = 2(\|h(x_i)-h(x_j)\|-m_p)\times sgn(h(x_j)-h(x_i))\times I_c \\
			& I_c = I_{m_p-\|h(x_i)-h(x_j)\|^2>0}
		\end{split}
	\end{equation}
	where $I_c$ is the indicator function as described above.
	
	These derivative values can be fed into the network via the back-propagation algorithm to update the parameters of each layer in the deep network. The training procedure is ended until the loss converges or a predefined maximal iteration number is reached. After the deep model is trained, we can generate \textit{q}-bit binary codes for any input image. For an input image $x$, it is first encoded into a hash codes $h(x) \in [0, 1]^q$ by the trained model. Then we can compute binary code for hashing by equation (\ref{binarycode}).
	
	In this way, the deep network can be trained end-to-end in a semi-supervised fashion, and newly input images can be encoded into binary hash codes by the deep model, we can perform efficient image retrieval via fast Hamming distance ranking between binary hash codes.
	\begin{figure*}[!htpb]
		\centering
		\includegraphics[width=0.8\textwidth]{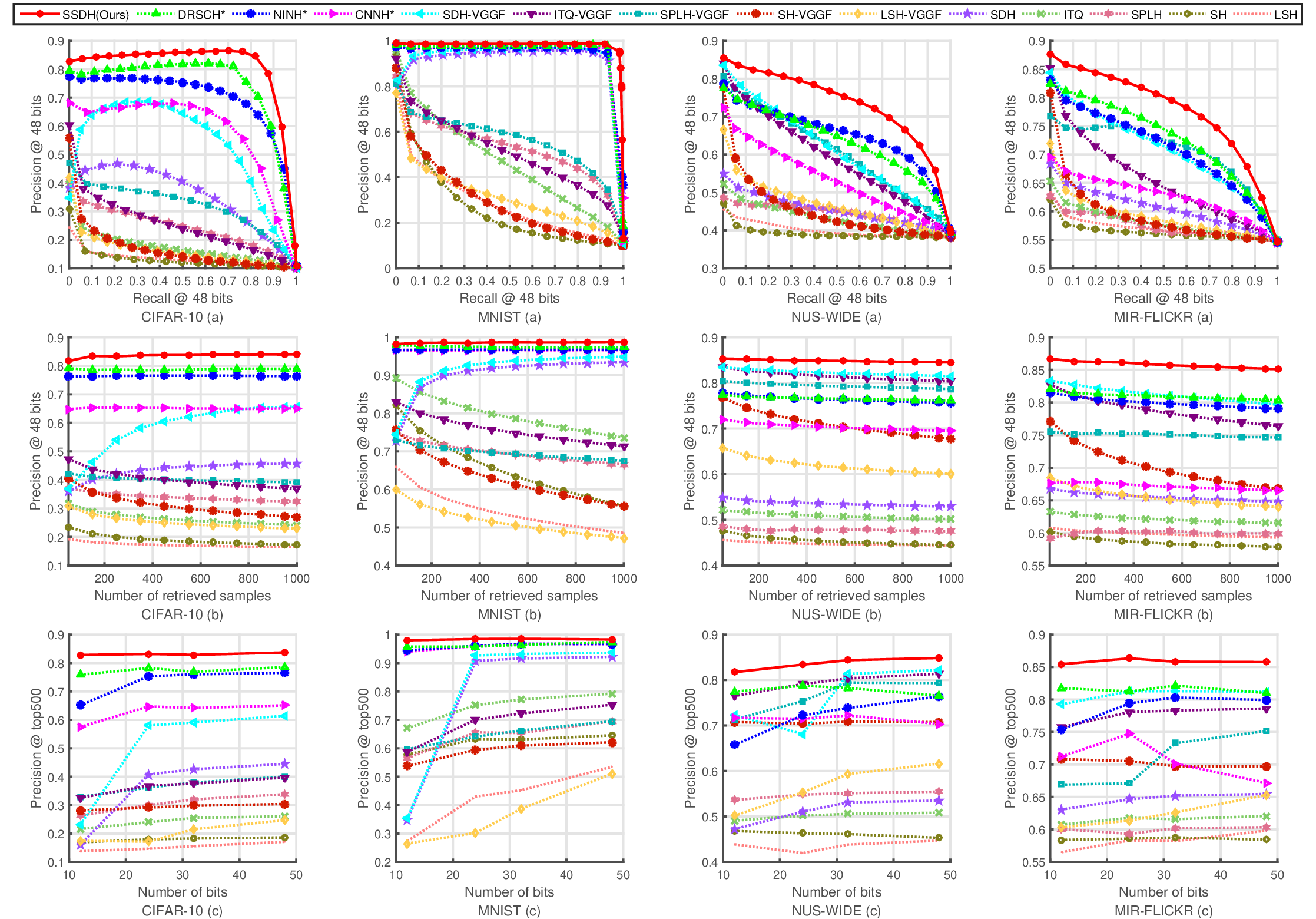}
		\caption{The comparison results on four datasets. The first row is the Precision-Recall curves with 48bit hash codes. The second row is the Precision@topk curves with 48bit hash codes. The third row is the Precision@top500 w.r.t. different length of hash codes.}
		\label{mnistfigure}
	\end{figure*}
	\section{Experiments}
	In this section, we first introduce the experiments conducted on 4 widely-used datasets (MNIST, CIFAR-10, NUS-WIDE and MIRFLICKR) with 8 state-of-the-art methods, including unsupervised methods LSH~\cite{gionis1999similarity}, SH~\cite{weiss2009spectral} and ITQ~\cite{gong2011iterative}, semi-supervised method SPLH~\cite{wang2010sequential}, supervised methods SDH~\cite{7298598}, CNNH~\cite{xia2014supervised}, NINH~\cite{7298947} and DRSCH~\cite{zhang2015bit}. LSH, SH, ITQ, SPLH and SDH are traditional hashing methods, which take hand-crafted image features as input to learn hash functions, while CNNH, NINH, DRSCH and our proposed SSDH approach are deep hashing methods, which take raw image pixels as input to conduct hash function learning. To further analyze the proposed SSDH approach, we conduct several baseline experiments to demonstrate the impact of each term in proposed loss function, as well as the impact of parameters $\lambda$ and $\mu$ and online graph construction. We further conduct experiments on a challenging transfer type testing protocols to comprehensively evaluate the proposed approach. Finally we evaluate on a large scale ImageNet dataset.
		\begin{table}[ht]
		\centering
		\begin{scriptsize}
			\caption{Split of each dataset}
			\label{Traditionalsetting}
			\begin{tabular}{c|c|c|c|c}
				\hline
				& MNIST & CIFAR10 & NUS-WIDE & MIRFLICKR  \\ \hline
				Query    & 1,000 & 1,000    & 2,100    & 1,000         \\ \hline
				Retrieval Database & 64,000 & 54,000   & 153,447    & 19,000        \\ \hline
				Labeled training set & 5,000 & 5,000   & 10,500    & 5,000         \\ \hline
			\end{tabular}
		\end{scriptsize}
	\end{table}
	\subsection{Datasets and evaluation metrics}
	We conduct experiments on 4 widely-used image datasets. Each dataset is split into query, retrieval database and labeled training set, where the labeled training set is randomly selected from retrieval database. We summarize the split of each dataset in table~\ref{Traditionalsetting}. Our proposed approach and compared state-of-the-art methods are divided into three categories: unsupervised, supervised and semi-supervised methods. Unsupervised methods use retrieval database as training set, supervised methods use the labeled data as training set, and semi-supervised methods including our proposed approach utilize both labeled data and unlabeled database as training set. The detailed descriptions and settings of each dataset are as follows:
	\begin{itemize}
		\item The \textbf{CIFAR-10}\footnote{http://www.cs.toronto.edu/~kriz/cifar.html} dataset consists of 60,000 color images with the resolution of $32\times 32$ from 10 categories, each of which has 6,000 images. For fair comparison, following~\cite{7298947,xia2014supervised}, 1,000 images are randomly selected as the query set (100 images per class), the rest images are used as retrieval database, and 5,000 images (500 images per class) are further randomly selected to form the labeled training set.
		\item The \textbf{MNIST}\footnote{http://yann.lecun.com/exdb/mnist/} dataset contains 70,000 gray scale handwritten images from ``0'' to ``9'' with the resolution of $28\times 28$. Following~\cite{xia2014supervised}, 1,000 images are randomly selected as the query set (100 images per class), the rest images are used as retrieval database, and 5,000 images (500 images per class) are randomly selected from the database as the labeled training set.
		\item The \textbf{NUS-WIDE}\footnote{http://lms.comp.nus.edu.sg/research/NUS-WIDE.htm}~\cite{chua2009nus} dataset contains nearly 270,000 images, and each image is associated with one or multiple labels of 81 semantic concepts. Following \cite{liu2011hashing,7298947}, only the 21 most frequent concepts are used, where each concept has at least 5,000 images, resulting in a total of 166,047 images. 2,100 images are randomly selected as the query set (100 images per concept), the rest images are used as retrieval database, and 500 images from each of the 21 concepts are randomly selected to form the labeled training set.
		\item The \textbf{MIRFLICKR}\footnote{http://press.liacs.nl/mirflickr/}~\cite{huiskes2008mir} dataset consists of 25,000 images collected from Flickr, and each image is associated with one or multiple labels of 38 semantic concepts. 1,000 images are randomly selected as the query set, the rest images are used as retrieval database, and 5,000 images are randomly selected from the rest of images to form the training set.
	\end{itemize}
	
	To objectively and comprehensively evaluate the retrieval accuracy of the proposed approach and all compared methods, we use four evaluation metrics: Mean Average Precision (MAP), Precision-Recall curves, Precision@topk and Precision@top500, which are defined as follows:
	\begin{itemize}
		\item Mean Average Precision (MAP): MAP presents an overall measurement of the retrieval performance. MAP for a set of queries is the mean of the average precision (AP) for each query, where AP is defined as:
		\begin{equation}
			\small
			AP=\frac{1}{R}\sum_{k=1}^{n}\frac{k}{R_k}\times rel_k
		\end{equation}
		where \textit{n} is the size of database set, \textit{R} is the number of relevant images in database set, $R_k$ is the number of relevant images in the top \textit{k} returns, and $rel_k=1$ if the image ranked at \textit{kth} position is relevant and 0 otherwise.
		\item Precision-Recall curves: The precisions at certain level of recall, we calculate PR curves of all returned results.
		\item Precision@topk: The average precision of top \textit{k} returned images for each query.
		\item Precision@top500: The average precision of the top 500 returned image for each query with different lengths of hash codes.
	\end{itemize}
	
	\subsection{Implementation details}
	\label{Implementation Details}
	We implement the proposed SSDH method based on the open-source framework Caffe~\cite{jia2014caffe}. The parameters of our network are initialized with the CNN-F network~\cite{chatfield2014return}, which is pre-trained on ImageNet~\cite{ILSVRC15}. Similar initialization strategy has been used in other deep hashing methods~\cite{zhao2015deep,zhu2016deep}. In all experiments, our networks are trained with the initial learning rate of 0.001, and we divide the learning rate by 10 each 20,000 steps. The size of mini-batch is 128. The weight decay parameter is 0.0005. For the parameters in the proposed loss function, we set $\lambda=0.1$ and $\mu=0.1$ in all experiments, we will further conduct experiments on the parameters to show that retrieval accuracy is not sensitive to parameter settings.
	
	For the proposed SSDH method, CNNH, NINH and DRSCH, we use raw image pixels as input. The implementations of CNNH~\cite{xia2014supervised} and DRSCH~\cite{zhang2015bit} are provided by their authors, while NINH~\cite{7298947} is of our own implementation. Since the network structures of these deep hashing methods are different, for a fair comparison, we use the \textit{same} CNN-F network as the base structure for the four deep methods. The network parameters of deep hashing methods are initialized with the \textit{same} pre-trained CNN-F model, thus we can perform fair comparison among these deep hashing methods. The results of CNNH~\cite{xia2014supervised}, NINH~\cite{7298947} and DRSCH~\cite{zhang2015bit} are referred as CNNH$\ast$, NINH$\ast$ and DRSCH$\ast$ respectively.
	
	For other compared methods without deep networks, i.e., the traditional hashing methods, we represent each image by hand-crafted features and deep features respectively. For hand-crafted features, we represent images in the \textbf{CIFAR-10} and \textbf{MIRFLICKR} by 512-dimensional GIST features, images in \textbf{MNIST} by 784-dimensional gray scale values, and images in the \textbf{NUS-WIDE} by 500-dimensional bag-of-words features. For a fair comparison between traditional methods and deep hashing methods, we also conduct experiments on the traditional methods with  features extracted from deep networks, where we extract 4096-dimensional deep feature from the activation of second fully-connected layer (fc7) of the pre-trained CNN-F network. We denote the results of traditional methods using hand-crafted features by LSH, SH, ITQ, SPLH and SDH, and we denote the results of traditional methods using deep features by LSH-CNNF, SH-CNNF, ITQ-CNNF, SPLH-CNNF and SDH-CNNF. The results of SDH, SH and ITQ are obtained from the implementations provided by the authors of original paper, while the results of LSH and SPLH are from our own implementation.
	
	\subsection{Experiment Results and Analysis}
	\begin{table*}[]
		\centering
		\begin{scriptsize}
			\caption{MAP scores with different length of hash codes on CIFAR10, MNIST, NUS-WIDE and MIRFLICKR dataset}
			\label{ResultTableALL}
			\begin{tabular}{c|cccc|cccc|cccc|cccc}
				\hline
				\multirow{2}{*}{MAP} & \multicolumn{4}{c|}{\textbf{CIFAR10}}  & \multicolumn{4}{c|}{\textbf{MNIST}}    & \multicolumn{4}{c|}{\textbf{NUS-WIDE}}  & \multicolumn{4}{c}{\textbf{MIRFLICKR}} \\ \cline{2-17} 
				& 12bit & 24bit & 32bit & 48bit & 12bit & 24bit & 32bit & 48bit & 12bit & 24bit & 32bit & 48bit & 12bit  & 24bit & 32bit & 48bit \\ \hline
				\textbf{SSDH(Ours)}                 & \textbf{0.801} & \textbf{0.813} &\textbf{ 0.812} & \textbf{0.814} & \textbf{0.975} & \textbf{0.982} & \textbf{0.982} & \textbf{0.982} & \textbf{0.707} & \textbf{0.725} & \textbf{0.731} & \textbf{0.735} & \textbf{0.773}  & \textbf{0.779} &\textbf{ 0.778} & \textbf{0.778 }\\ \hline
				DRSCH$\ast$               & 0.721 & 0.733 & 0.726 & 0.747 & 0.951 & 0.953 & 0.955 & 0.966 & 0.640 & 0.650 & 0.655 & 0.635 & 0.741  & 0.741 & 0.737 & 0.728 \\ \hline
				NINH$\ast$                & 0.600 & 0.696 & 0.689 & 0.702 & 0.931 & 0.949 & 0.958 & 0.959 & 0.597 & 0.627 & 0.647 & 0.651 & 0.693  & 0.711 & 0.718 & 0.709 \\ \hline
				CNNH$\ast$                & 0.496 & 0.580 & 0.582 & 0.583 & 0.925 & 0.955 & 0.964 & 0.965 & 0.536 & 0.522 & 0.533 & 0.531 & 0.667  & 0.688 & 0.654 & 0.626 \\ \hline
				SDH-CNNF             & 0.363 & 0.528 & 0.529 & 0.542 & 0.542 & 0.938 & 0.943 & 0.944 & 0.520 & 0.507 & 0.591 & 0.610 & 0.695  & 0.704 & 0.697 & 0.708 \\ \hline
				SPLH-CNNF            & 0.268 & 0.282 & 0.299 & 0.318 & 0.493 & 0.517 & 0.526 & 0.559 & 0.582 & 0.594 & 0.612 & 0.607 & 0.648  & 0.659 & 0.695 & 0.706 \\ \hline
				ITQ-CNNF             & 0.219 & 0.228 & 0.239 & 0.247 & 0.407 & 0.478 & 0.487 & 0.506 & 0.582 & 0.581 & 0.583 & 0.588 & 0.648  & 0.654 & 0.652 & 0.652 \\ \hline
				SH-CNNF              & 0.169 & 0.161 & 0.161 & 0.159 & 0.301 & 0.304 & 0.296 & 0.287 & 0.486 & 0.462 & 0.455 & 0.448 & 0.603  & 0.595 & 0.590 & 0.588 \\ \hline
				LSH-CNNF             & 0.132 & 0.124 & 0.144 & 0.157 & 0.176 & 0.191 & 0.220 & 0.305 & 0.432 & 0.451 & 0.464 & 0.466 & 0.571  & 0.574 & 0.580 & 0.589 \\ \hline
				SDH        & 0.255 & 0.330 & 0.344 & 0.360 & 0.526 & 0.915 & 0.921 & 0.926 & 0.414 & 0.465 & 0.451 & 0.454 & 0.595  & 0.601 & 0.608 & 0.605 \\ \hline
				SPLH       & 0.209 & 0.227 & 0.237 & 0.244 & 0.455 & 0.494 & 0.513 & 0.526 & 0.415 & 0.420 & 0.431 & 0.440 & 0.575  & 0.574 & 0.575 & 0.574 \\ \hline
				ITQ        & 0.158 & 0.163 & 0.168 & 0.169 & 0.404 & 0.442 & 0.447 & 0.460 & 0.428 & 0.429 & 0.430 & 0.431 & 0.576  & 0.579 & 0.579 & 0.580 \\ \hline
				SH         & 0.124 & 0.125 & 0.125 & 0.126 & 0.290 & 0.278 & 0.260 & 0.254 & 0.390 & 0.391 & 0.389 & 0.390 & 0.561  & 0.562 & 0.563 & 0.562 \\ \hline
				LSH        & 0.116 & 0.121 & 0.124 & 0.131 & 0.162 & 0.236 & 0.222 & 0.276 & 0.404 & 0.384 & 0.394 & 0.400 & 0.557  & 0.564 & 0.562 & 0.569 \\ \hline
			\end{tabular}
		\end{scriptsize}
	\end{table*}
	Table~\ref{ResultTableALL} shows the MAP scores of all the retrieved results on CIFAR10, MNIST, NUS-WIDE and MIRFLICKR datasets with different length of hash codes. We can observe that: 
	(1) The proposed SSDH approach outperforms compared state-of-the-art supervised deep hashing methods. For example on CIFAR10 dataset, the proposed SSDH approach improves the average MAP from 73.2\% (DRSCH$\ast$), 67.2\% (NINH$\ast$) and 56.0\% (CNNH$\ast$) to 81.0\%. Similar results can be observed on MNIST, NUS-WIDE and MIRFLICKR dataset. It's because the proposed SSDH approach designs a semi-supervised loss function, which not only preserves the semantic similarity provided by labels, but also captures the meaningful neighborhood structure, thus improves the search accuracy.
	(2) SSDH, DSRCH$\ast$ and NINH$\ast$ significantly outperform the two-stage method CNNH$\ast$, because they learn hash codes and image features simultaneously, which can benefit from each other, thus leading to better performance compared to CNNH$\ast$.
	(3) The trends among traditional methods are that supervised methods SDH outperform semi-supervised method SPLH, which further outperforms unsupervised method.	We can also observe that the traditional hashing methods using deep features significantly outperform the identical methods using hand-crafted features. For example on CIFAR10, compared to SDH, SDH-CNNF improves the average MAP score from 32.2\% to 49.0\%, which indicates that deep features can also boosts traditional hashing methods' performance.
	
	The first row of Fig.~\ref{mnistfigure} shows the Precision-Recall curves with 48bit hash codes on four datasets. We can clearly see that the proposed SSDH approach outperforms other state-of-the-art hashing methods, which demonstrates the effectiveness of exploiting the underlying structures of unlabeled data.
	The second row of Fig.~\ref{mnistfigure} demonstrates the Precision@topk curves with 48bit hash codes, it can be observed that SSDH achieves the best search accuracy steadily when the number of retrieved results increases, which has superior advantage over state-of-the-art methods.
	The third row of Fig.~\ref{mnistfigure} shows the Precision@top500 for different code lengths on four datasets. We can clearly see that the proposed SSDH approach outperforms all compared state-of-the-art methods on all hash code length. For example on CIFAR10 dataset, the proposed SSDH approach achieves over 80\% search accuracy on all hash code lengths. Also from all these figures, by comparing traditional methods using deep features and the identical methods using hand-crafted features, we can observer that deep features can boost the performance of traditional methods.
	
	In all metrics on the 4 widely-used datasets, the proposed SSDH approach shows superior advantages over the state-of-the-art hashing methods, which demonstrates the effectiveness of proposed SSDH method. SSDH outperforms supervised deep hashing methods NINH$\ast$, DRSCH$\ast$ and CNNH$\ast$ in all evaluation metrics, which indicates that the proposed SSDH method can preserve the semantic similarity and underlying data structures simultaneously, thus leading to better retrieval accuracy.
	
	\subsection{Comparison of Computation Time}
	Besides the retrieval performance, we also compare the computation time cost of the proposed SSDH approach with other compared methods. All the experiments are carried out on the same PC with NVIDIA Titan X GPU, Intel Core i7-5930k 3.50GHz CPU and 64 GB memory. Hashing based image retrieval process generally consists of three parts: Feature extraction, hash codes generation and image search among database. Note that our proposed SSDH approach and other deep hashing methods are end-to-end frameworks whose inputs are raw images, and outputs are hash codes. For a fair comparison between deep hashing methods and traditional methods, we compare with traditional methods by deep features input. We record feature extraction time, hash codes generation time and image search time cost of each method. The average computation time of different methods are shown in table~\ref{efficiency}. We can observe that, among deep hashing methods, DRSCH is relatively slow, because it uses weighted Hamming distance to perform image retrieval, which is slower than original Hamming distance. The computation time of our proposed SSDH approach and other 6 compared methods is comparable with each other, because when hash functions are learned, the time cost of hash generation is only a matrix multiplication which is very fast, and all of them use Hamming distance that can be efficiently implemented by bit-wise XOR operation.
	
	\begin{table}[htb]
		\centering
		\caption{Comparison of the average computation time (Millisecond per Image) on four benchmark dataset by fixing the code length 48.}
		\label{efficiency}
		\begin{tabularx}{0.5\textwidth}{c|YYYY}
			\hline
			Method   & CIFAR10 \newline (ms) & MNIST\newline(ms) & NUSWIDE\newline(ms) & MIRFLICKR\newline(ms) \\ \hline
			\textbf{SSDH(ours)}     & \textbf{3.43}   & \textbf{4.31}  & \textbf{11.56}     & \textbf{2.12}      \\ \hline
			DRSCH$\ast$    & 4.44    & 4.92  & 12.56     & 2.83      \\ \hline
			NINH$\ast$     & 3.92    & 4.46  & 11.94     & 2.26      \\ \hline
			CNNH$\ast$     & 3.78    & 4.14  & 11.92     & 2.34      \\ \hline
			SDH-CNNF  & 4.24    & 4.35  & 12.02     & 2.33      \\ \hline
			SPLH-CNNF & 4.05    & 4.43  & 12.31     & 2.39      \\ \hline
			ITQ-CNNF  & 4.22    & 4.67  & 12.24     & 2.50      \\ \hline
			SH-CNNF   & 4.41    & 4.55  & 12.07     & 2.32      \\ \hline
			LSH-CNNF  & 4.15    & 4.65  & 12.28     & 2.30      \\ \hline
		\end{tabularx}
	\end{table}
	\subsection{Baseline Experiments and Analysis}
	
	\subsubsection{Evaluation of the Loss Function}
\begin{figure}[htb]
		\centering
		\includegraphics[width=0.4\textwidth]{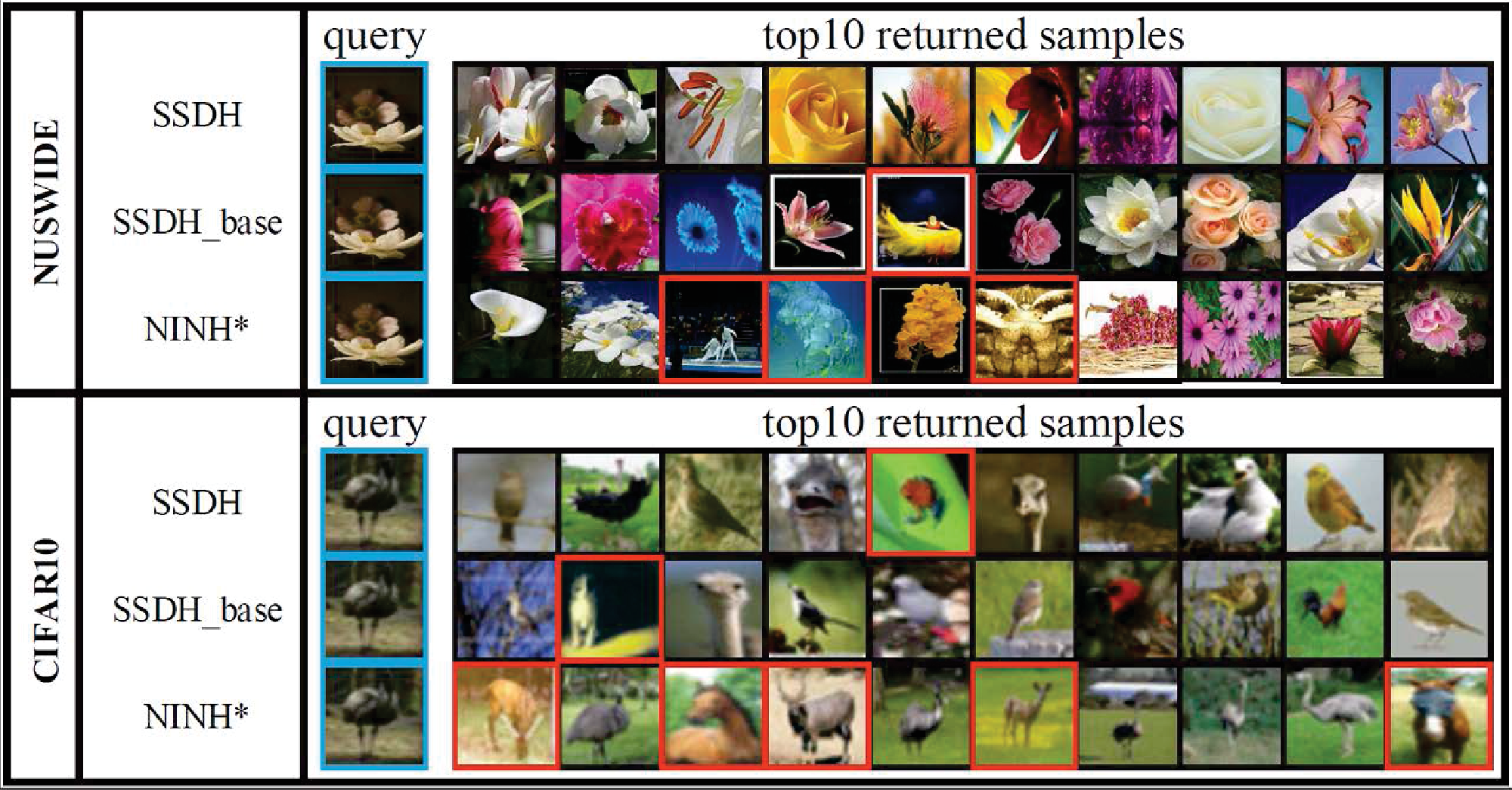}
		\caption{Some retrieval results of NUS-WIDE and CIFAR10 using Hamming ranking on 48bit hash codes. The blue rectangles denote the query images. The red rectangles indicate wrong retrieval results. We can observe that SSDH achieves the best results, and SSDH\_base achieves better results than NINH$\ast$.}
		\label{demofig}
	\end{figure}

\begin{figure*}[htb]
		\centering
		\includegraphics[width=0.8\textwidth]{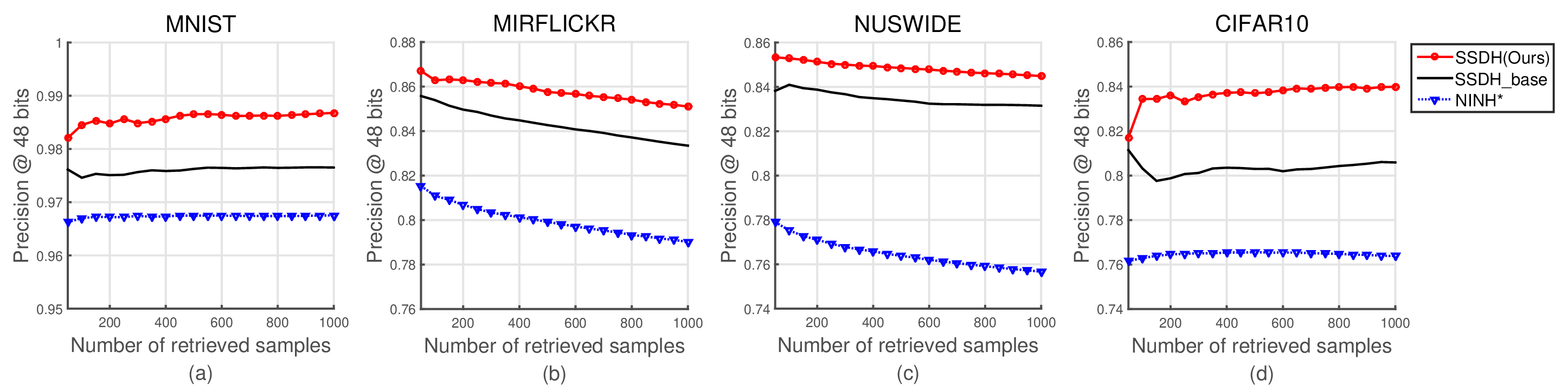}
		\caption{Precision@topk curves with 48bit hash codes on 4 datasets. (a) MNIST; (b) MIRFLICKR; (c) NUS-WIDE; (d) CIFAR10.}
		\label{baseprecision}
	\end{figure*}

\begin{figure*}[htb]
		\centering
		\includegraphics[width=0.8\textwidth]{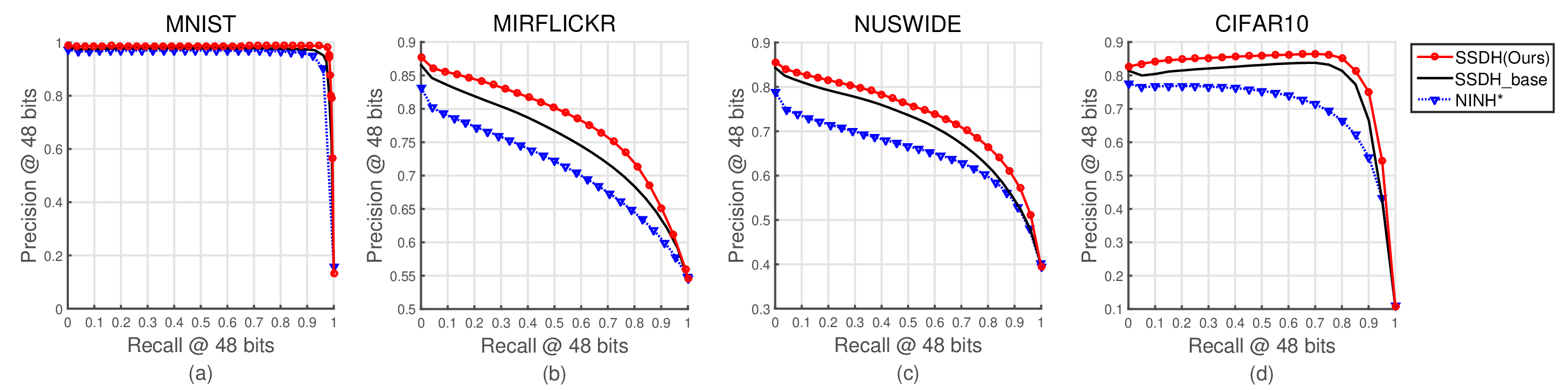}
		\caption{Precision-Recall curves of Hamming ranking with 48bit hash codes on 4 datasets. (a) MNIST; (b) MIRFLICKR; (c) NUS-WIDE; (d) CIFAR10.}
		\label{basepr}
	\end{figure*}

\begin{figure*}[htb]
		\centering
		\includegraphics[width=0.8\textwidth]{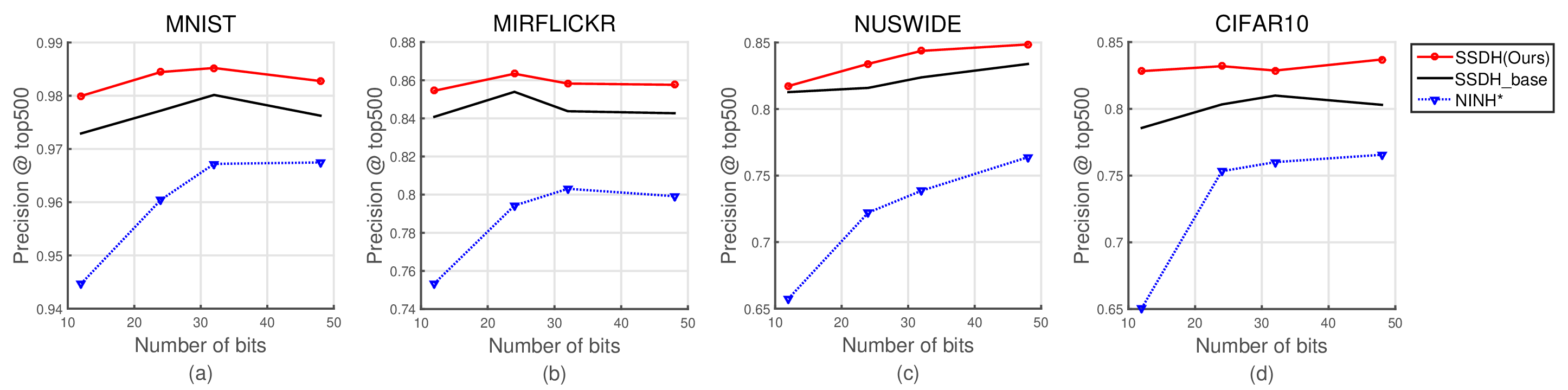}
		\caption{Precision@top500 w.r.t. different length of hash codes on 4 datasets. (a) MNIST; (b) MIRFLICKR; (c) NUS-WIDE; (d) CIFAR10.}
		\label{baseprat500}
	\end{figure*}

	To further verify the effectiveness of proposed semi-supervised loss function, we also report results of using first supervised ranking term only (NINH$\ast$), and results of using supervised ranking term and semi-supervised embedding term (SSDH\_base). By comparing SSDH\_base with NINH$\ast$, we can verify the effectiveness of semi-supervised embedding term. By comparing SSDH with SSDH\_base, we can verify the complementarity of semi-supervised embedding term and pseudo-label term. The MAP scores are shown in table~\ref{losstable}. SSDH\_base achieves better results on all 4 datasets with all hash code lengths than NINH$\ast$, which demonstrates the effectiveness of semi-supervised embedding term. SSDH approach outperforms SSDH\_base on 4 datasets and all hash code lengths, which verifies the complementarity of semi-supervised embedding term and pseudo-label term. Fig.~\ref{baseprecision} demonstrates the Precision@topk curves with 48bit hash codes on 4 datasets. Fig.~\ref{basepr} shows the Precision-Recall curves of Hamming ranking with 48bit hash codes on 4 datasets. Fig.~\ref{baseprat500} shows the Precision@top500 w.r.t. different length of hash codes on 4 datasets. From these three figures, we can clearly observe that SSDH outperforms SSDH\_base and SSDH\_base outperforms NINH$\ast$, which further demonstrates the effectiveness of each term in our proposed loss function. Fig.~\ref{demofig} demonstrates the top 10 retrieval results of NUS-WIDE and CIFAR10 using Hamming ranking on 48bit hash codes.
	
	\begin{table*}[htb]
		\centering
		\caption{Evaluation of three terms in the proposed loss function on 4 datasets.}
		\label{losstable}
		\begin{tabularx}{0.9\textwidth}{c|YYYY|YYYY|YYYY}
			\hline
			\multirow{2}{*}{Methods} & \multicolumn{4}{c|}{SSDH (All three terms)}     & \multicolumn{4}{c|}{SSDH\_base (First \& second term)} & \multicolumn{4}{c}{NINH$\ast$ (First term only)}    \\ \cline{2-13} 
			& 12bit & 24bit & 32bit & 48bit & 12bit  & 24bit  & 32bit & 48bit & 12bit & 24bit & 32bit & 48bit \\ \hline
			MNIST                    & 0.975 & 0.982 & 0.982 & 0.982 & 0.944  & 0.969  & 0.975 & 0.969 & 0.931 & 0.949 & 0.958 & 0.959 \\ \hline
			CIFAR10                  & 0.801 & 0.813 & 0.812 & 0.814 & 0.749  & 0.773  & 0.784 & 0.779 & 0.600 & 0.696 & 0.689 & 0.702 \\ \hline
			NUS-WIDE                 & 0.707 & 0.725 & 0.731 & 0.735 & 0.665  & 0.672  & 0.691 & 0.706 & 0.597 & 0.627 & 0.647 & 0.651 \\ \hline
			MIRFLICKR                & 0.773 & 0.779 & 0.778 & 0.778 & 0.747  & 0.758  & 0.759 & 0.751 & 0.693 & 0.711 & 0.718 & 0.709 \\ \hline
		\end{tabularx}
	\end{table*}
	
	\subsubsection{Evaluation of Parameters}
\begin{figure}[!htpb]
		\centering
		\includegraphics[width=0.4\textwidth]{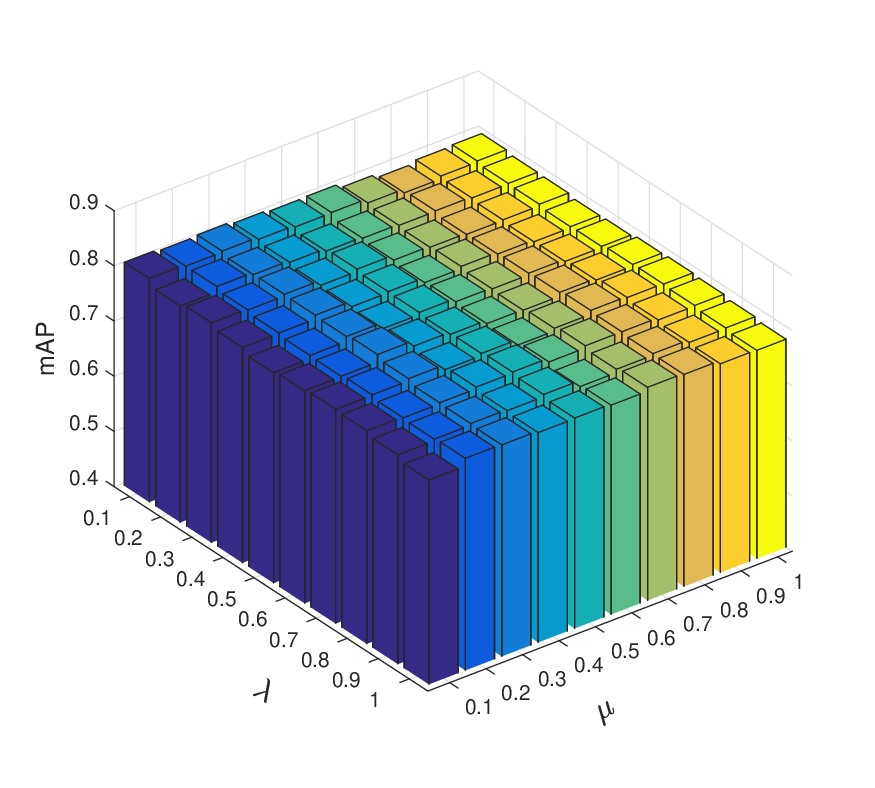}
		\caption{MAP scores with respect to parameter variations on CIFAR10 dataset with 48bit code length.}
		\label{params}
	\end{figure}

	The proposed SSDH approach has two parameters $\lambda$ and $\mu$, which control the impact of semi-supervised embedding term and pseudo-label term. We set them with different values ranging from 0.1 to 1, and compute MAP scores on CIFAR10 dataset with 48bit code length. The results are shown in Fig.~\ref{params}, and it can be observed that experimental results are not sensitive to these parameters.
	\subsubsection{Evaluation of Online Graph Construction}
	To further demonstrate the effectiveness of our proposed online graph construction strategy, we conduct a baseline experiment to compare with offline graph construction strategy. More specifically, we first construct a neighborhood graph offline based on all the data. It is constructed based on deep features extracted from the fc7 layer of the pre-trained CNN-F network. Then we use the proposed SSDH approach to train a deep hashing network based on the offline neighborhood graph. The only difference between offline and online approach is that in equation (\ref{jointloss}): offline approach randomly selects neighbor pairs and non-neighbor pairs based on the fixed offline graph, while online approach is based on the evolving graph constructed online. We conduct this baseline experiment on CIFAR10 dataset, the results are shown in table~\ref{graph construction}. We can observe that, online graph construction achieves better retrieval results than offline graph construction on all hash code length, which shows that online graph construction can benefit from the evolving deep features and capture more meaningful neighbors. We can also observe that offline graph construction can improve the retrieval accuracy compared with state-of-the-art methods in table~\ref{ResultTableALL}, which also demonstrates the effectiveness of simultaneously preserving the semantic similarity and the underlying data structures in our proposed SSDH approach.
	
	\begin{table}[htb]
		\centering
		\caption{MAP scores of online and offline graph construction strategy on CIFAR10 dataset}
		\label{graph construction}
		\begin{tabularx}{0.49\textwidth}{c|YYYY}
			\hline
			\multirow{2}{*}{Methods} & \multicolumn{4}{c}{CIFAR10}  \\ \cline{2-5} 
			& 12bit & 24bit & 32bit & 48bit \\ \hline
			SSDH with Online Graph             & 0.801 & 0.813 & 0.812 & 0.814 \\ \hline
			SSDH with Offline Graph            & 0.756 & 0.764 & 0.756 & 0.766 \\ \hline
		\end{tabularx}
	\end{table}
	
\begin{figure}[htb]
		\centering
		\includegraphics[width=0.4\textwidth]{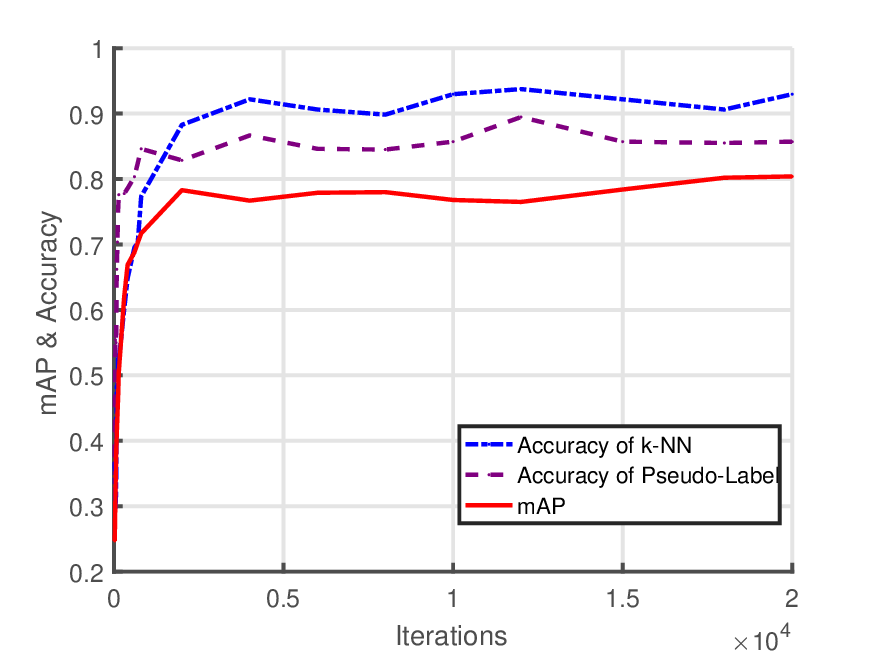}
		\caption{The accuracy of neighborhood graph and predicted pseudo labels, as well as the MAP scores of image retrieval during each iteration on CIFAR10 dataset with 48bit code length.}
		\label{knngraph}
	\end{figure}

	To further demonstrate the representative power of neighborhood graph and pseudo labels, we also conduct experiments to analyze the accuracy of the online constructed graph and predicted pseudo labels. More specifically, during the training stage we calculate the accuracy of neighbors among the constructed graph and the accuracy of predicted pseudo labels of unlabeled data. We also calculate MAP score of image retrieval by using the model of corresponding iterations, so that we can clearly see the relations between them. The results are shown in Fig.~\ref{knngraph}. We can observe that both the accuracies of neighborhood graph and predicted pseudo labels increase fast during training, and achieve over 80\% within 2,000 iterations, which indicates that the online graph and the predicted pseudo labels are becoming more and more accurate due to the evolving deep features. From Fig.~\ref{knngraph} we can also observe that the accuracy of image retrieval increases along with the neighborhood graph and pseudo labels, which indicates that proposed SSDH captures more and more meaningful semantic neighbors and improves the search accuracy.	

	\subsubsection{Evaluation of the Impact of k}
\begin{figure}[htb]
		\centering
		\includegraphics[width=0.4\textwidth]{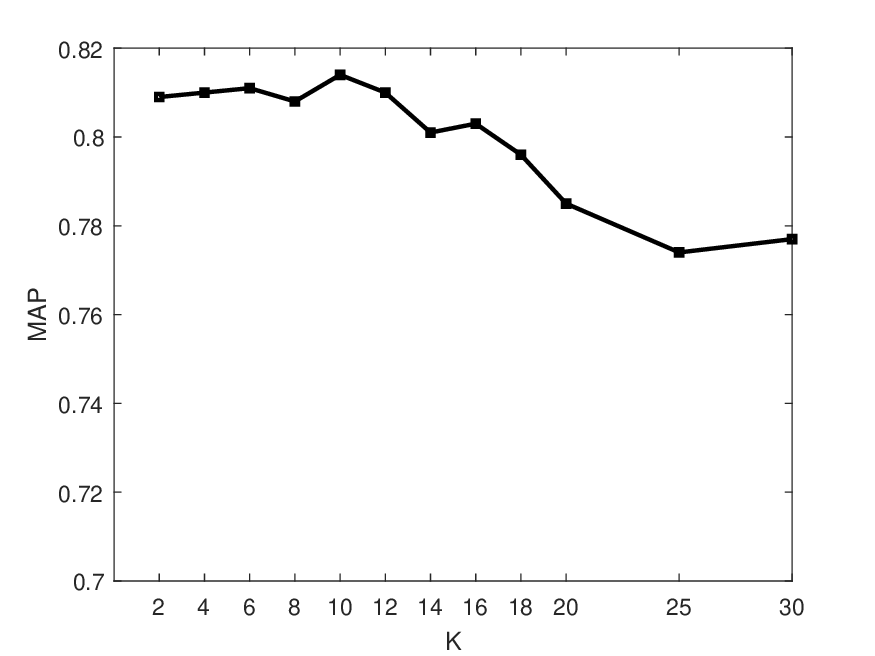}
		\caption{The MAP score w.r.t different k on CIFAR10 dataset.}
		\label{impactofk}
	\end{figure}

	We conduct experiments to study the impact of \textit{k} in \textit{k}-nearest-neighbor graph. More specifically, we calculate the MAP score of image retrieval w.r.t different \textit{k} ranging from 2 to 30. We conduct this baseline experiment on CIFAR10 dataset with 48 bit code lenght. The result is shown in Fig.~\ref{impactofk}, from which we can observe that the MAP score is stable when $k<10$, but it decreases when $k>10$. This trend is expected, and it is related to the accuracy of \textit{k}-nearest-neighbors. The accuracy of \textit{k}-nearest-neighbors will decrease when \textit{k} is large, which will further lead to low retrieval accuracy. However, the result is stable when $k<10$, which can ensure the robustness of our proposed online graph construction approach.

	\subsection{Experiment Results on Transfer Type Testing Protocols}
	Besides the traditional testing protocols used by most hashing works, we also conduct experiments under transfer type testing protocols proposed in~\cite{sablayrolles2016should}, where only 75\% of categories are known during training, and the remaining 25\% of categories are used for evaluation. We follow the exactly same setting, and randomly split each dataset into 4 sets, namely train75, test75, train25 and test25, where train75 and test75 are the images of 75\% categories, while train25 and test25 are the images of the remaining 25\% categories. The training set is formed by train75, the query set is formed by test25, and the database set is formed by train25 and test75. Note that test75 is a distraction set in database, since there are no ground truth images in them. The detailed split of each dataset under transfer type protocol is shown in table~\ref{Transfersetting}. The experiment results are shown in table~\ref{transfertable}. For fair comparison, we report traditional methods by only using deep features. From table~\ref{transfertable}, we can observe that the overall retrieval accuracy of each method decreases under transfer type testing protocol, which is because of the information loss of unseen categories. The retrieval accuracy gap between supervised methods and unsupervised methods becomes smaller than that of traditional settings. For example, ITQ even achieves better retrieval accuracy than SDH and CNNH on NUS-WIDE dataset. It is because unsupervised methods don't depend on the semantic information, which are more robust under transfer type testing protocols, while supervised methods are likely to overfitting to the known categories. Traditional semi-supervised method SPLH achieves promising results, which even outperforms deep hashing methods on some datasets, it is because semi-supervised paradigm can make use of labeled data as well as prevent overfitting. Under the transfer type testing protocols, our proposed SSDH approach still achieves the best results, because proposed SSDH preserves the semantic similarity and the underlying data structures of unlabeled data simultaneously, thus it's more robust to the unseen categories and achieves the best results.
	
	\begin{table}[htb]
		\centering
		\caption{Split of each dataset under transfer type testing protocol}
		\label{Transfersetting}
		\begin{tabular}{c|cccc}
			\hline
			& CIFAR10 & MNIST & NUS-WIDE & MIRFLICKR \\ \hline
			Query    & 9000 &  10913    & 35176    & 6976   \\ \hline
			Database & 30000 & 35000   & 83023    & 12709    \\ \hline
			Training & 21000 & 24088   & 47848    & 5315     \\ \hline
		\end{tabular}
	\end{table}
	
	\begin{table*}[htb]
		\centering
		\begin{scriptsize}
			\caption{MAP scores with different length of hash codes on CIFAR10, MNIST, NUS-WIDE and MIRFLICKR dataset under transfer type testing protocol}
			\label{transfertable}
			\begin{tabular}{c|cccc|cccc|cccc|cccc}
				\hline
				\multirow{2}{*}{Methods} & \multicolumn{4}{c|}{\textbf{CIFAR10}}  & \multicolumn{4}{c|}{\textbf{MNIST}}    & \multicolumn{4}{c|}{\textbf{NUS-WIDE}} & \multicolumn{4}{c}{\textbf{MIRFLICKR}} \\ \cline{2-17} 
				& 12bit & 24bit & 32bit & 48bit & 12bit & 24bit & 32bit & 48bit & 12bit & 24bit & 32bit & 48bit & 12bit  & 24bit & 32bit & 48bit \\ \hline
				\textbf{SSDH(Ours)}                 & \textbf{0.274} & \textbf{0.286} & \textbf{0.305} & \textbf{0.304} & \textbf{0.344} &\textbf{0.357} & \textbf{0.404} & \textbf{0.477} & \textbf{0.513} & \textbf{0.536} & \textbf{0.541} & \textbf{0.547} & \textbf{0.780}  & \textbf{0.783} & \textbf{0.781} & \textbf{0.785} \\ \hline
				DRSCH$\ast$               & 0.239 & 0.223 & 0.240 & 0.231 & 0.316 & 0.321 & 0.321 & 0.335 & 0.454 & 0.467 & 0.465 & 0.456 & 0.756  & 0.754 & 0.755 & 0.761 \\ \hline
				NINH$\ast$                & 0.235 & 0.248 & 0.251 & 0.269 & 0.313 & 0.330 & 0.373 & 0.394 & 0.484 & 0.483 & 0.485 & 0.487 & 0.751  & 0.755 & 0.760 & 0.761 \\ \hline
				CNNH$\ast$                & 0.229 & 0.235 & 0.221 & 0.216 & 0.266 & 0.273 & 0.260 & 0.281 & 0.455 & 0.473 & 0.478 & 0.478 & 0.740  & 0.742 & 0.742 & 0.737 \\ \hline
				SDH-CNNF             & 0.179 & 0.190 & 0.196 & 0.204 & 0.225 & 0.268 & 0.305 & 0.317 & 0.469 & 0.505 & 0.497 & 0.517 & 0.692  & 0.706 & 0.720 & 0.732 \\ \hline
				SPLH-CNNF            & 0.200 & 0.228 & 0.231 & 0.222 & 0.282 & 0.296 & 0.294 & 0.293 & 0.458 & 0.464 & 0.521 & 0.524 & 0.705  & 0.709 & 0.712 & 0.744 \\ \hline
				ITQ-CNNF             & 0.147 & 0.159 & 0.173 & 0.180 & 0.283 & 0.308 & 0.326 & 0.345 & 0.488 & 0.486 & 0.498 & 0.493 & 0.683  & 0.683 & 0.684 & 0.685 \\ \hline
				SH-CNNF              & 0.127 & 0.117 & 0.115 & 0.112 & 0.227 & 0.244 & 0.247 & 0.241 & 0.396 & 0.372 & 0.361 & 0.356 & 0.623  & 0.614 & 0.615 & 0.611 \\ \hline
				LSH-CNNF             & 0.078 & 0.098 & 0.105 & 0.096 & 0.186 & 0.288 & 0.247 & 0.244 & 0.310 & 0.323 & 0.336 & 0.343 & 0.591  & 0.609 & 0.611 & 0.616 \\ \hline
			\end{tabular}
		\end{scriptsize}
	\end{table*}
	\subsection{Experiment Results on Large Scale ImageNet Dataset}
	The \textbf{ImageNet} dataset (ILSVRC12)~\cite{ILSVRC15} contains 1000 categories with 1.2 million images. ImageNet is a large scale dataset which can comprehensively evaluate the effectiveness of the proposed approach and compared state-of-the-art methods. In order to fairly evaluate the retrieval accuracy on the ImageNet dataset, we no longer use the pre-trained CNN-F model to initialize the parameters of the network structure, since the pre-trained model has learned the information from all the 1.2 million training images. Thus we also use the transfer type testing protocols on the ImageNet dataset. More specifically, we randomly split ImageNet dataset into 4 sets, namely train75, test75, train25 and test25, where train75 and test75 are the images of 750 categories, while train25 and test25 are the images of the remaining 250 categories. Similar to the above setting, we use train75 as training set, test25 as query set, and the union of test75 and train25 as database set. Thus we construct a training set of 499,071 images, query set of 166,513 images and database set of 665,583 images. For a fair comparison between deep hashing methods and traditional methods, we first use the training set to train the CNN-F model from scratch, then the trained model is used to initialize the parameters of network structure for deep hashing methods, and to extract 4096 dimensional features for traditional hashing methods. For the training of each hashing methods, since the number of images in training set is too large to train the compared hashing methods, we further randomly select 75,000 images (100 images per category) to train hashing methods. 
	
	For this large-scale dataset, we only report the precisions within top 500 retrieved samples with different hash code length due to high computation cost of MAP evaluation. The results are shown in table~\ref{imagenetresult}. Since this dataset is very large and we test on more challenging transfer type testing protocols, the precision values are relatively low. From table~\ref{imagenetresult} we can observe that our proposed SSDH approach still achieves the best results on this large scale ImageNet dataset and challenging testing protocol, which demonstrates the effectiveness of our proposed SSDH approach.
	
	\begin{table}[htb]
		\centering
		\caption{The retrieval precisions within top 500 retrieved samples w.r.t. different hash code length on ImageNet dataset}
		\label{imagenetresult}
		\begin{tabularx}{0.4\textwidth}{c|YYYY}
			\hline
			\multirow{2}{*}{Methods} & \multicolumn{4}{c}{\textbf{ImageNet}} \\ \cline{2-5} 
			& 12bit & 24bit & 32bit & 48bit \\ \hline
			\textbf{SSDH(ours)}                     & \textbf{0.084} & \textbf{0.106} & \textbf{0.122} &\textbf{ 0.134} \\ \hline
			DRSCH$\ast$                   & 0.063 & 0.089 & 0.101 & 0.106 \\ \hline
			NINH$\ast$                      & 0.062 & 0.090 & 0.095 & 0.105 \\ \hline
			CNNH$\ast$                      & 0.020 & 0.041 & 0.056 & 0.081 \\ \hline
			SDH-CNNF                 & 0.037 & 0.070 & 0.086 & 0.105 \\ \hline
			SPLH-CNNF                & 0.053 & 0.080 & 0.090 & 0.103 \\ \hline
			ITQ-CNNF                 & 0.044 & 0.074 & 0.091 & 0.108 \\ \hline
			SH-CNNF                  & 0.040 & 0.064 & 0.075 & 0.087 \\ \hline
			LSH-CNNF                 & 0.010 & 0.015 & 0.018 & 0.023 \\ \hline
		\end{tabularx}
	\end{table}
	
	\section{Conclusion}
	In this paper, we have proposed a novel semi-supervised deep hashing (SSDH) method. Firstly, we design a deep network with an online graph construction strategy to make full use of unlabeled data efficiently, which can perform hash code learning and image feature learning simultaneously in a semi-supervised fashion. Secondly, we propose a semi-supervised loss function with a supervised ranking term to minimize the empirical error on labeled data, as well as a semi-supervised embedding term and a pseudo-label term to minimize the embedding error on both labeled and unlabeled data, which can capture the semantic similarity information and the underlying structures of data. Experimental results show the effectiveness of our method compared with 8 state-of-the-art methods. 
	
	For the future work, we will explore different semi-supervised embedding algorithms that can make better use of unlabeled data, and more advanced graph construction strategies can be utilized. We also plan to extend this framework to an unsupervised scenario, in which we intend to use clustering methods to obtain virtual labels of images.
	\ifCLASSOPTIONcaptionsoff
	\newpage
	\fi

	
	
	\bibliographystyle{IEEEtran}
	\bibliography{ssdh}

\begin{thebibliography}{10}
\providecommand{\url}[1]{#1}
\csname url@samestyle\endcsname
\providecommand{\newblock}{\relax}
\providecommand{\bibinfo}[2]{#2}
\providecommand{\BIBentrySTDinterwordspacing}{\spaceskip=0pt\relax}
\providecommand{\BIBentryALTinterwordstretchfactor}{4}
\providecommand{\BIBentryALTinterwordspacing}{\spaceskip=\fontdimen2\font plus
\BIBentryALTinterwordstretchfactor\fontdimen3\font minus
  \fontdimen4\font\relax}
\providecommand{\BIBforeignlanguage}[2]{{%
\expandafter\ifx\csname l@#1\endcsname\relax
\typeout{** WARNING: IEEEtran.bst: No hyphenation pattern has been}%
\typeout{** loaded for the language `#1'. Using the pattern for}%
\typeout{** the default language instead.}%
\else
\language=\csname l@#1\endcsname
\fi
#2}}
\providecommand{\BIBdecl}{\relax}
\BIBdecl

\bibitem{imghashsurvey}
J.~Wang, T.~Zhang, s.~j, N.~Sebe, and H.~T. Shen, ``A survey on learning to
  hash,'' \emph{IEEE Transactions on Pattern Analysis and Machine Intelligence
  (PAMI)}, vol.~PP, no.~99, pp. 1--1, 2017.

\bibitem{gong2011iterative}
Y.~Gong and S.~Lazebnik, ``Iterative quantization: A procrustean approach to
  learning binary codes,'' in \emph{IEEE Conference on Computer Vision and
  Pattern Recognition (CVPR)}, 2011, pp. 817--824.

\bibitem{irie2014locally}
G.~Irie, Z.~Li, X.-M. Wu, and S.-F. Chang, ``Locally linear hashing for
  extracting non-linear manifolds,'' in \emph{IEEE Conference on Computer
  Vision and Pattern Recognition (CVPR)}, 2014, pp. 2115--2122.

\bibitem{kan2014semisupervised}
M.~Kan, D.~Xu, S.~Shan, and X.~Chen, ``Semisupervised hashing via kernel
  hyperplane learning for scalable image search,'' \emph{IEEE Transactions on
  Circuits and Systems for Video Technology (TCSVT)}, vol.~24, no.~4, pp.
  704--713, 2014.

\bibitem{liu2011hashing}
W.~Liu, J.~Wang, S.~Kumar, and S.-F. Chang, ``Hashing with graphs,'' in
  \emph{International Conference on Machine Learning (ICML)}, 2011, pp. 1--8.

\bibitem{weiss2009spectral}
Y.~Weiss, A.~Torralba, and R.~Fergus, ``Spectral hashing,'' in \emph{Advances
  in neural information processing systems (NIPS)}, 2009, pp. 1753--1760.

\bibitem{7298947}
H.~Lai, Y.~Pan, Y.~Liu, and S.~Yan, ``Simultaneous feature learning and hash
  coding with deep neural networks,'' in \emph{IEEE Conference on Computer
  Vision and Pattern Recognition (CVPR)}, 2015, pp. 3270--3278.

\bibitem{xia2014supervised}
R.~Xia, Y.~Pan, H.~Lai, C.~Liu, and S.~Yan, ``Supervised hashing for image
  retrieval via image representation learning,'' in \emph{AAAI Conference on
  Artificial Intelligence (AAAI)}, 2014, pp. 2156--2162.

\bibitem{zhao2015deep}
F.~Zhao, Y.~Huang, L.~Wang, and T.~Tan, ``Deep semantic ranking based hashing
  for multi-label image retrieval,'' in \emph{IEEE Conference on Computer
  Vision and Pattern Recognition (CVPR)}, 2015, pp. 1556--1564.

\bibitem{zhu2016deep}
H.~Zhu, M.~Long, J.~Wang, and Y.~Cao, ``Deep hashing network for efficient
  similarity retrieval,'' in \emph{AAAI Conference on Artificial Intelligence
  (AAAI)}, 2016, pp. 2415--2421.

\bibitem{7937842}
S.~Zhang, J.~Li, M.~Jiang, and B.~Zhang, ``Scalable discrete supervised
  multimedia hash learning with clustering,'' \emph{IEEE Transactions on
  Circuits and Systems for Video Technology (TCSVT)}, vol.~PP, no.~99, pp.
  1--1, 2017.

\bibitem{7855682}
Z.~Chen, J.~Lu, J.~Feng, and J.~Zhou, ``Nonlinear structural hashing for
  scalable video search,'' \emph{IEEE Transactions on Circuits and Systems for
  Video Technology (TCSVT)}, vol.~PP, no.~99, pp. 1--1, 2017.

\bibitem{7851077}
A.~Araujo and B.~Girod, ``Large-scale video retrieval using image queries,''
  \emph{IEEE Transactions on Circuits and Systems for Video Technology
  (TCSVT)}, vol.~PP, no.~99, pp. 1--1, 2017.

\bibitem{gionis1999similarity}
A.~Gionis, P.~Indyk, R.~Motwani \emph{et~al.}, ``Similarity search in high
  dimensions via hashing,'' in \emph{International Conference on Very Large
  Data Bases (VLDB)}, vol.~99, no.~6, 1999, pp. 518--529.

\bibitem{wang2010sequential}
J.~Wang, S.~Kumar, and S.-F. Chang, ``Sequential projection learning for
  hashing with compact codes,'' in \emph{International Conference on Machine
  Learning (ICML)}, 2010, pp. 1127--1134.

\bibitem{zhang2015bit}
R.~Zhang, L.~Lin, R.~Zhang, W.~Zuo, and L.~Zhang, ``Bit-scalable deep hashing
  with regularized similarity learning for image retrieval and person
  re-identification,'' \emph{IEEE Transactions on Image Processing (TIP)},
  vol.~24, no.~12, pp. 4766--4779, 2015.

\bibitem{Liu_2016_CVPR}
H.~Liu, R.~Wang, S.~Shan, and X.~Chen, ``Deep supervised hashing for fast image
  retrieval,'' in \emph{IEEE Conference on Computer Vision and Pattern
  Recognition (CVPR)}, 2016, pp. 2064--2072.

\bibitem{zhang2013topology}
L.~Zhang, Y.~Zhang, J.~Tang, X.~Gu, J.~Li, and Q.~Tian, ``Topology preserving
  hashing for similarity search,'' in \emph{ACM International Conference on
  Multimedia (ACM-MM)}, 2013, pp. 123--132.

\bibitem{AIBC}
F.~Shen, Y.~Yang, L.~Liu, W.~Liu, D.~Tao, and H.~T. Shen, ``Asymmetric binary
  coding for image search,'' \emph{IEEE Transactions on Multimedia (TMM)},
  vol.~19, no.~9, pp. 2022--2032, 2017.

\bibitem{kulis2009learning}
B.~Kulis and T.~Darrell, ``Learning to hash with binary reconstructive
  embeddings,'' in \emph{Advances in neural information processing systems
  (NIPS)}, 2009, pp. 1042--1050.

\bibitem{kulis2009fast}
B.~Kulis, P.~Jain, and K.~Grauman, ``Fast similarity search for learned
  metrics,'' \emph{IEEE Transactions on Pattern Analysis and Machine
  Intelligence (PAMI)}, vol.~31, no.~12, pp. 2143--2157, 2009.

\bibitem{norouzi2011minimal}
M.~Norouzi and D.~M. Blei, ``Minimal loss hashing for compact binary codes,''
  in \emph{International Conference on Machine Learning (ICML)}, 2011, pp.
  353--360.

\bibitem{liu2012supervised}
W.~Liu, J.~Wang, R.~Ji, Y.-G. Jiang, and S.-F. Chang, ``Supervised hashing with
  kernels,'' in \emph{IEEE Conference on Computer Vision and Pattern
  Recognition (CVPR)}, 2012, pp. 2074--2081.

\bibitem{norouzi2012hamming}
M.~Norouzi, D.~J. Fleet, and R.~R. Salakhutdinov, ``Hamming distance metric
  learning,'' in \emph{Advances in neural information processing systems
  (NIPS)}, 2012, pp. 1061--1069.

\bibitem{wang2013learning}
J.~Wang, W.~Liu, A.~X. Sun, and Y.-G. Jiang, ``Learning hash codes with
  listwise supervision,'' in \emph{IEEE International Conference on Computer
  Vision (ICCV)}, 2013, pp. 3032--3039.

\bibitem{wang2013order}
J.~Wang, J.~Wang, N.~Yu, and S.~Li, ``Order preserving hashing for approximate
  nearest neighbor search,'' in \emph{ACM International Conference on
  Multimedia (ACM-MM)}, 2013, pp. 133--142.

\bibitem{li2013learning}
X.~Li, G.~Lin, C.~Shen, A.~Van Den~Hengel, and A.~R. Dick, ``Learning hash
  functions using column generation.'' in \emph{International Conference on
  Machine Learning (ICML)}, 2013, pp. 142--150.

\bibitem{lin2014fast}
G.~Lin, C.~Shen, Q.~Shi, A.~van~den Hengel, and D.~Suter, ``Fast supervised
  hashing with decision trees for high-dimensional data,'' in \emph{IEEE
  Conference on Computer Vision and Pattern Recognition (CVPR)}, 2014, pp.
  1963--1970.

\bibitem{wang2015ranking}
Q.~Wang, Z.~Zhang, and L.~Si, ``Ranking preserving hashing for fast similarity
  search,'' in \emph{International Joint Conference on Artificial Intelligence
  (IJCAI)}, 2015, pp. 3911--3917.

\bibitem{7298598}
F.~Shen, C.~Shen, W.~Liu, and H.~Tao~Shen, ``Supervised discrete hashing,'' in
  \emph{IEEE Conference on Computer Vision and Pattern Recognition (CVPR)},
  2015, pp. 37--45.

\bibitem{DPLM}
F.~Shen, X.~Zhou, Y.~Yang, J.~Song, H.~T. Shen, and D.~Tao, ``A fast
  optimization method for general binary code learning,'' \emph{IEEE
  Transactions on Image Processing (TIP)}, vol.~25, no.~12, pp. 5610--5621,
  2016.

\bibitem{oliva2001modeling}
A.~Oliva and A.~Torralba, ``Modeling the shape of the scene: A holistic
  representation of the spatial envelope,'' \emph{International Journal of
  Computer Vision (IJCV)}, vol.~42, no.~3, pp. 145--175, 2001.

\bibitem{krizhevsky2012imagenet}
A.~Krizhevsky, I.~Sutskever, and G.~E. Hinton, ``Imagenet classification with
  deep convolutional neural networks,'' in \emph{Advances in neural information
  processing systems (NIPS)}, 2012, pp. 1097--1105.

\bibitem{hinton2012improving}
G.~E. Hinton, N.~Srivastava, A.~Krizhevsky, I.~Sutskever, and R.~R.
  Salakhutdinov, ``Improving neural networks by preventing co-adaptation of
  feature detectors,'' \emph{arXiv preprint arXiv:1207.0580}, 2012.

\bibitem{lin2013network}
M.~Lin, Q.~Chen, and S.~Yan, ``Network in network,'' \emph{arXiv preprint
  arXiv:1312.4400}, 2013.

\bibitem{chatfield2014return}
K.~Chatfield, K.~Simonyan, A.~Vedaldi, and A.~Zisserman, ``Return of the devil
  in the details: Delving deep into convolutional nets,'' \emph{arXiv preprint
  arXiv:1405.3531}, 2014.

\bibitem{chapelle2006semi}
O.~Chapelle, B.~Sch{\"o}lkopf, A.~Zien \emph{et~al.}, ``Semi-supervised
  learning,'' \emph{MIT press Cambridge}, 2006.

\bibitem{hadsell2006dimensionality}
R.~Hadsell, S.~Chopra, and Y.~LeCun, ``Dimensionality reduction by learning an
  invariant mapping,'' in \emph{IEEE Conference on Computer Vision and Pattern
  Recognition (CVPR)}, 2006, pp. 1735--1742.

\bibitem{lee2013pseudo}
D.-H. Lee, ``Pseudo-label: The simple and efficient semi-supervised learning
  method for deep neural networks,'' in \emph{Workshop on Challenges in
  Representation Learning, ICML}, 2013, pp. 2--7.

\bibitem{MLKNN}
M.-L. Zhang and Z.-H. Zhou, ``Ml-knn: A lazy learning approach to multi-label
  learning,'' \emph{Pattern recognition (PR)}, vol.~40, no.~7, pp. 2038--2048,
  2007.

\bibitem{chua2009nus}
T.-S. Chua, J.~Tang, R.~Hong, H.~Li, Z.~Luo, and Y.~Zheng, ``Nus-wide: a
  real-world web image database from national university of singapore,'' in
  \emph{ACM international conference on image and video retrieval (CIVR)},
  2009, pp. 1--9.

\bibitem{huiskes2008mir}
M.~J. Huiskes and M.~S. Lew, ``The mir flickr retrieval evaluation,'' in
  \emph{ACM international conference on Multimedia information retrieval
  (MIR)}, 2008, pp. 39--43.

\bibitem{jia2014caffe}
Y.~Jia, E.~Shelhamer, J.~Donahue, S.~Karayev, J.~Long, R.~Girshick,
  S.~Guadarrama, and T.~Darrell, ``Caffe: Convolutional architecture for fast
  feature embedding,'' in \emph{ACM International Conference on Multimedia
  (ACM-MM)}, 2014, pp. 675--678.

\bibitem{ILSVRC15}
O.~Russakovsky, J.~Deng, H.~Su, J.~Krause, S.~Satheesh, S.~Ma, Z.~Huang,
  A.~Karpathy, A.~Khosla, M.~Bernstein \emph{et~al.}, ``Imagenet large scale
  visual recognition challenge,'' \emph{International Journal of Computer
  Vision (IJCV)}, vol. 115, no.~3, pp. 211--252, 2015.

\bibitem{sablayrolles2016should}
A.~Sablayrolles, M.~Douze, H.~J{\'e}gou, and N.~Usunier, ``How should we
  evaluate supervised hashing?'' \emph{arXiv preprint arXiv:1609.06753}, 2016.

\end{thebibliography}
	%
	
	
	
	%
	
	\begin{IEEEbiography}[{\includegraphics[width=1in,height=1.25in,clip,keepaspectratio]{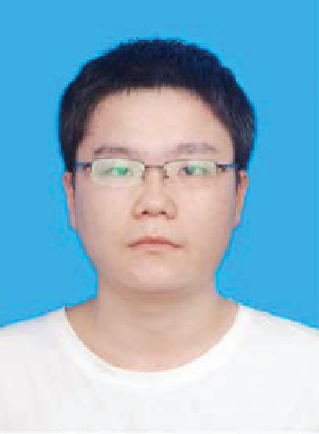}}] {Jian Zhang}
		received the B.S. degree in computer science and technology from Peking University, in Jul. 2012. He is currently pursuing the Ph.D. degree	in the Institute of Computer Science and Technology
		(ICST), Peking University. His research interests include multimedia retrieval and machine learning.
	\end{IEEEbiography}
	\begin{IEEEbiography}[{\includegraphics[width=1in,height=1.25in,clip,keepaspectratio]{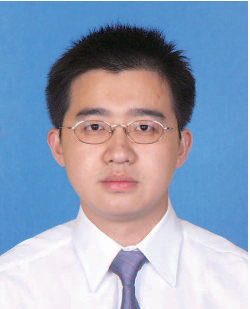}}] {Yuxin Peng}
	is currently the professor of Institute of Computer Science and Technology (ICST), Peking University, Beijing, China, and the chief scientist of 863 Program (National Hi-Tech Research and Development Program of China). He received the Ph.D. degree in computer application from Peking University, in July 2003. After that, he worked as an assistant professor in ICST, Peking University. He was promoted to an associate professor and professor in Peking University in August 2005 and August 2010, respectively. In 2006, he was authorized by the “Program for New Star in Science and Technology of Beijing” and the “Program for New Century Excellent Talents in University (NCET).” He has published more than 100 papers in refereed international journals and conference proceedings, including IJCV, TIP, TCSVT, TMM, PR, ACM MM, ICCV, CVPR, IJCAI, AAAI, etc. He led his team to participate in TRECVID (TREC Video Retrieval Evaluation) many times. In TRECVID 2009, his team won four first places on 4 sub-tasks of the High-Level Feature Extraction task and Search task. In TRECVID 2012, his team gained four first places on all 4 sub-tasks of the Instance Search (INS) task and Known-Item Search task. In TRECVID 2014, his team gained the first place in the Interactive Instance Search task. His team also gained both two first places in the INS task of TRECVID 2015, 2016, and 2017. Besides, he won the first prize of Beijing Science and Technology Award for Technological Invention in 2016 (ranking first). He has applied 35 patents, and obtained 16 of them. His current research interests mainly include cross-media analysis and reasoning, image and video analysis and retrieval, and computer vision.
	\end{IEEEbiography}
	
	
	
	
	
	
	

\end{document}